\documentclass[11pt]{article}
\usepackage[preprint]{acl}
\usepackage{times}
\usepackage{latexsym}
\usepackage{inconsolata}
\usepackage[utf8]{inputenc}
\usepackage[T1]{fontenc}
\usepackage{graphicx}
\usepackage{stfloats}

\usepackage{hyperref}
\hypersetup{colorlinks=true,citecolor=darkblue,linkcolor=darkblue,urlcolor=darkblue}
\usepackage{url}
\usepackage{booktabs}\usepackage{multirow}
\usepackage{amsmath}
\usepackage{amssymb}
\usepackage{amsfonts}
\usepackage{microtype}
\usepackage{xcolor}
\usepackage{array}
\usepackage{tabularx}
\usepackage{longtable}
\usepackage{ragged2e}
\usepackage{pifont}
\usepackage{multirow}
\usepackage{colortbl}
\usepackage{algorithm}
\usepackage{algorithmic}
\usepackage{placeins}\usepackage{float}
\usepackage{needspace}
\usepackage{enumitem}
\newcommand{\VideoHALO}{\textsc{VideoHALO}}
\newcommand{\VidHalLoc}{\textsc{VidHalLoc}}
\definecolor{vidhalpink}{RGB}{255,240,245}
\definecolor{vidhalored}{RGB}{178,45,45}
\definecolor{answergreen}{RGB}{48,116,45}
\definecolor{answerred}{RGB}{190,45,35}
\definecolor{vidhalfamgray}{RGB}{244,244,244}
\definecolor{taskgreen}{RGB}{229,237,222}
\definecolor{taskblue}{RGB}{226,238,244}
\definecolor{taskyellow}{RGB}{255,244,218}
\newcommand{\accell}[2]{\cellcolor{vidhalored!#1}#2}
\newcommand{\accelllast}[2]{\multicolumn{1}{>{\columncolor{vidhalored!#1}[\tabcolsep][0pt]}c@{}}{#2}}

\newcommand{\taskband}[3]{%
  \begingroup\setlength{\fboxsep}{4pt}%
  \noindent\colorbox{#1}{\parbox{\dimexpr\linewidth-2\fboxsep\relax}{%
    \raggedright\ttfamily\fontsize{8.2}{9.7}\selectfont{\fontfamily{pcr}\fontseries{b}\selectfont #2}\par #3}}\par
  \endgroup\vspace{1.5pt}%
}
\newcommand{\taskplain}[2]{%
  \begingroup\raggedright\ttfamily\fontsize{8.2}{9.7}\selectfont
  \noindent{\fontfamily{pcr}\fontseries{b}\selectfont #1}\par #2\par\endgroup\vspace{1.5pt}%
}
\newcommand{\taskitem}[2]{%
  \noindent\hangindent=1.45em\hangafter=1\makebox[1.45em][r]{#1.\enspace}#2\par
}
\newcommand{\taskbullet}[1]{%
  \noindent\hangindent=1.25em\hangafter=1\makebox[1.25em][r]{\textbullet\enspace}#1\par
}
\newcommand{\taskfield}[2]{\taskbullet{{\fontfamily{pcr}\fontseries{b}\selectfont #1:} #2}}
\newcolumntype{Q}[1]{>{\centering\arraybackslash}p{#1}}
\newcommand{\yesmark}{\raisebox{-0.15ex}{\normalsize\ding{51}}}
\newcommand{\goodmark}{\raisebox{-0.18ex}{\includegraphics[height=1.40ex]{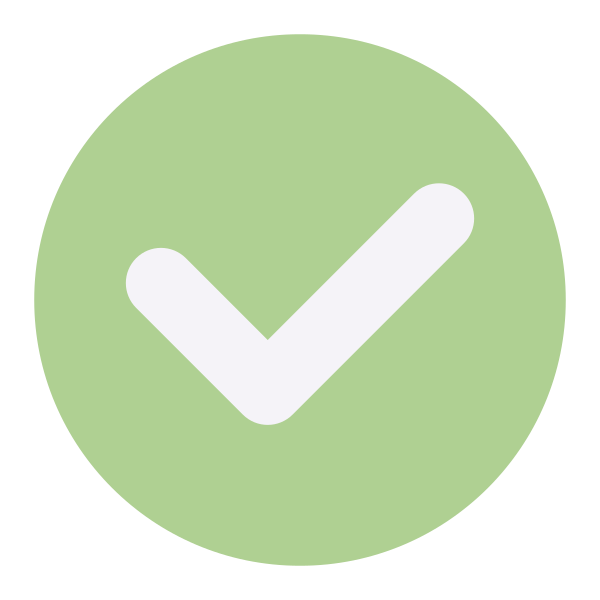}}}
\newcommand{\badmark}{\raisebox{-0.18ex}{\includegraphics[height=1.40ex]{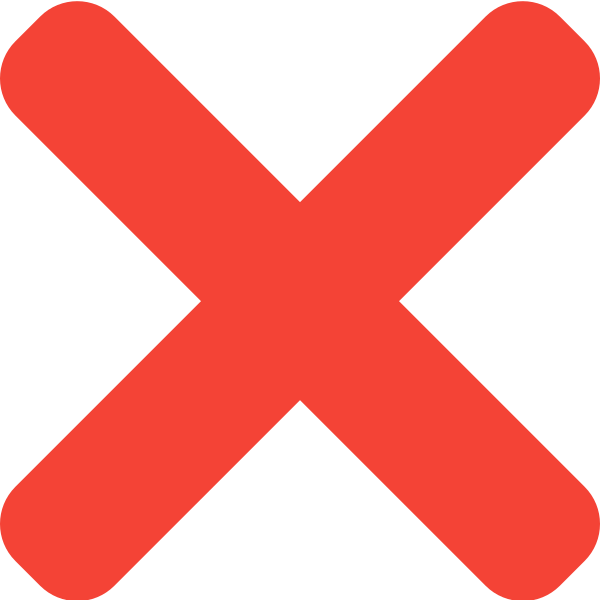}}}
\newcommand{\examplejudgeprompt}{You are judging whether one candidate answer is fully supported by a video and correctly answers the question. \textbf{Question:} \texttt{\{question\}}. \textbf{Candidate answer:} \texttt{\{candidate\_answer\}}. Judge this candidate independently using only the video, the question, and this candidate answer. Do not assume that another candidate answer exists, and do not compare this answer with any alternative answer. Return exactly one label and nothing else. Return \textsc{Accept} only when the candidate answer is fully supported by the video and correctly answers the question. Return \textsc{Reject} when the candidate answer is contradicted by the video, unsupported, incomplete, irrelevant, or otherwise does not correctly answer the question. If the available video evidence is insufficient to fully support the candidate answer, return \textsc{Reject}.}
\newcommand{\methodfamilyrow}[1]{%
  \hline
  \rowcolor{vidhalfamgray}[0pt][0pt]%
  \multicolumn{7}{@{}c@{}}{\rule[-0.62ex]{0pt}{2.65ex}\textcolor{black}{\itshape #1}} \\
  \hline
}
\title{Can We Trust Video Hallucination Detectors?\\\VidHalLoc{} for Evaluating the Evaluators}
\author{Xinyu CHEN \quad Adnan Mahmood \quad Mark Dras}
\DeclareRobustCommand{\equaldash}[1]{\leavevmode\unskip\nobreak\hspace{2pt}#1\nobreak\hspace{2pt}\ignorespaces}
\begin{document}
\flushbottom
\setlength{\parskip}{0pt}

\maketitle

\begin{abstract}
Video-language models and video agents can produce hallucinations that conflict with spatiotemporal evidence. Existing benchmarks mainly evaluate model hallucinations, and heterogeneous mechanisms make detector reliability difficult to compare. We introduce \textbf{\VidHalLoc{}}, a benchmark that evaluates hallucination detection methods under a unified diagnostic evaluation protocol using 2,000 adversarial hallucination samples across Video Question Answering and Video Captioning tasks, spanning Ontology and Dynamic hallucination categories. To construct \VidHalLoc{} efficiently, we introduce \textbf{\VideoHALO{}}, a Harness Engineering-informed multi-agent workflow that decomposes data construction into four executable stages supported by a memory system and a communication protocol. Evaluation of fifteen methods reveals that the four dedicated detectors peak at an Overall accuracy of only 34.63\%, indicating limited reliability across video hallucination types [\href{https://huggingface.co/datasets/wesfggfd/VidHalLoc}{Dataset Repository}].
\end{abstract}

\section{Introduction}\label{sec:introduction}
Large Video-Language Models (LVLMs) and video agents can now process long recordings, reason over complex spatial and temporal events, and coordinate tools for video understanding~\cite{google2025gemini3flash,openai2025gpt5,li2024llavanextinterleave,bai2025qwen3vl,gemmateam2026gemma4,wang2025internvl35,qwenteam2026qwen36,xu2025qwen3omni,fan2024videoagentmemory,wang2026think,zhang2025deepvideodiscovery}. These capabilities do not prevent them from producing content that is inconsistent with the video~\cite{zheng2025lvlms}. We define a \textit{video hallucination} as generated content that misaligns with observable video evidence. Compared with image hallucination, the video setting requires evidence to be traced across entity-related information, actions, temporal relations, and camera transitions. Reliable hallucination detection is therefore an essential part of trustworthy video-language systems.

Multimodal hallucination detection has evolved from image-level verification to video-aware assessment. Existing approaches assess video–text consistency through embedding similarity, learned entailment, and structured verification. Related approaches also estimate grounding confidence from internal model signals~\cite{shalam2026propose,jing2024faithscore,chen2024unified,bansal2024videocon,jing2025fifa}.

Image and video hallucination benchmarks have supported the study of trustworthy vision-language models by covering several forms of hallucinated content~\cite{li2023evaluating,wang2024amberllmfreemultidimensionalbenchmark,wang2024videohallucerevaluatingintrinsicextrinsic,li2025vidhalluc}. Most existing benchmarks focus on determining whether an LVLM or agent exhibits hallucinations, rather than evaluating the reliability of the hallucination detection methods themselves. The heterogeneity in their categorization systems and evaluation targets precludes meaningful comparisons across different detection paradigms. Furthermore, constructing a video benchmark with fine-grained type annotations poses a significant challenge, as evidence localization and manual verification demand considerable human involvement~\cite{wang2024videohallucerevaluatingintrinsicextrinsic,yang2024vript,zhang2025eventhallusiondiagnosingeventhallucinations,li2025vidhalluc,li2025videohallu,lu2025elv,lei2021qvhighlights,xiao2021nextqa,wang2024videocot}.

\begin{figure*}[!t]
\centering
\includegraphics[width=0.88\textwidth]{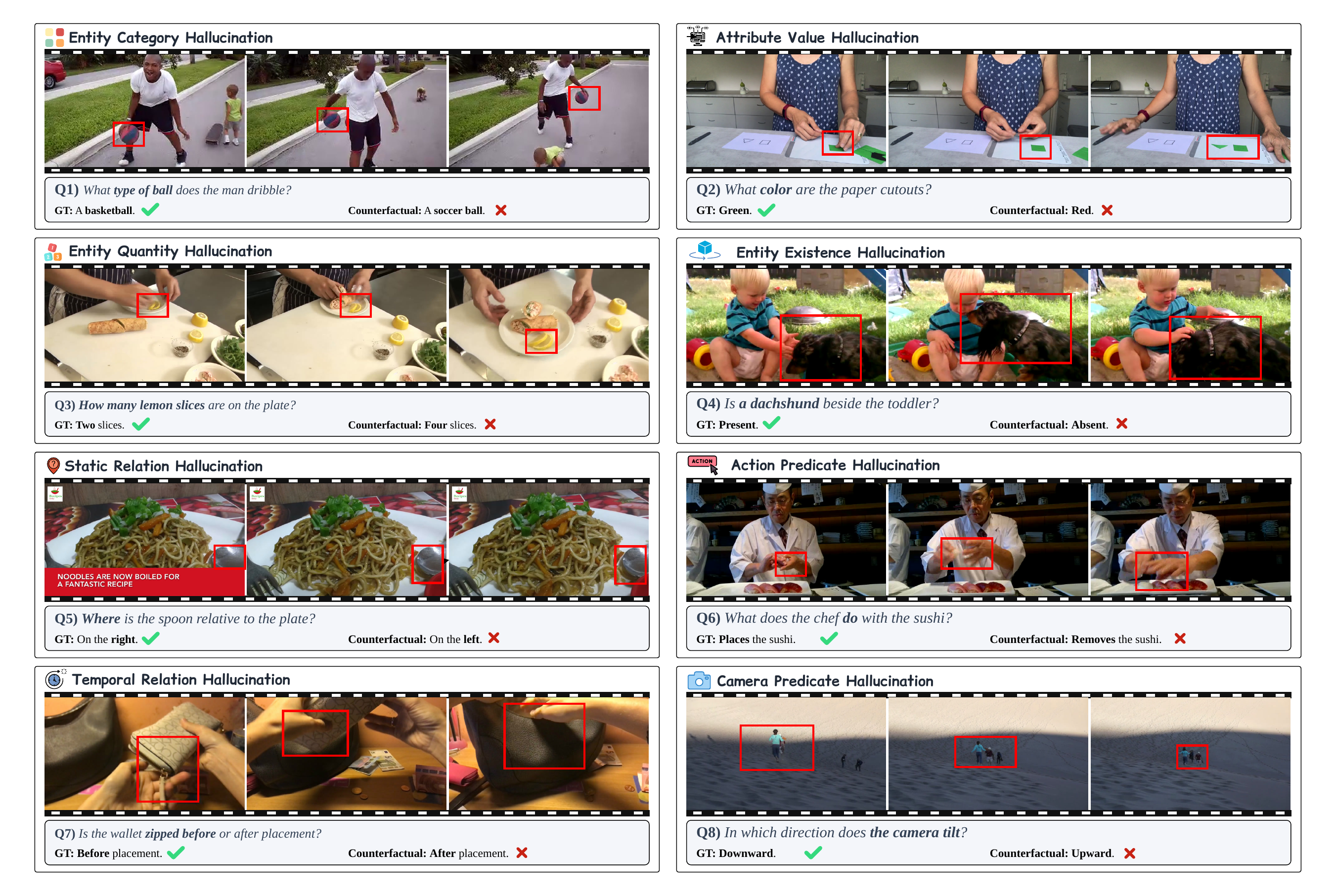}
\caption{An Overview Taxonomy of \VidHalLoc{}\kern-0.025em\equaldash{---}\kern-0.025em\VidHalLoc{} establishes a more comprehensive, hierarchical classification for the evaluation of hallucination detectors. It categorizes video hallucinations into ontology hallucinations (\textit{covering entity existence, category, quantity, attribute value, and static relations}) and dynamic hallucinations (\textit{encompassing action predicates, temporal relations, and camera predicates}). }
\label{fig:teaser}
\end{figure*}

To address this evaluation gap, we introduce \VidHalLoc{}, a benchmark for evaluating video hallucination detectors under a unified diagnostic framework and protocol (Figure~\ref{fig:teaser}). The benchmark is constructed using \VideoHALO{}, a multi-agent workflow informed by harness engineering~\cite{zhong2026aiharnessengineering,hong2024metagpt,wu2024autogen,qian2024chatdev,sumers2024coala}. The workflow achieves higher construction throughput than human experts with high category accuracy in the human audit (Table~\ref{tab:vidhalloc_efficiency}; Appendix~\ref{app:human_verification}). Among the fifteen evaluated methods, Gemini-3-Flash~\cite{google2025gemini3flash} achieves 83.63\% Overall accuracy, while the best dedicated detector reaches 34.63\%, indicating limited detector reliability across the evaluated video tasks.

\begin{table*}[!t]
\captionsetup{type=table}
\centering
\fontsize{8}{9}\selectfont
\renewcommand{\arraystretch}{1.00}
\setlength{\tabcolsep}{1.8pt}
\resizebox{\textwidth}{!}{%
\begin{tabular}{@{}lcccc@{\hspace{8pt}}cc@{}}
\toprule
\multirow[c]{2}{*}[-2.6pt]{\mbox{\textbf{Method}}} & \multirow[c]{2}{*}[-2.6pt]{\mbox{\textbf{Granularity}}} & \multirow[c]{2}{*}[-2.6pt]{\mbox{\textbf{Type Classification}}} & \multirow[c]{2}{*}[-2.6pt]{\mbox{\textbf{Detection}}} & \multicolumn{2}{c}{\mbox{\textbf{Output}}} & \multirow[c]{2}{*}[-2.6pt]{\mbox{\textbf{VLM Tasks}}} \\
\cmidrule(l{1pt}r{1pt}){5-6}
& & & & \mbox{\textbf{Score}} & \mbox{\textbf{Label}} & \\
\midrule
\mbox{\textbf{PAC-S} \mbox{\cite{sarto2023positive}}} & \mbox{Response} & \mbox{None} & \mbox{Trained Model} & \yesmark &  & \mbox{Img Cap., Vid. Caption} \\
\mbox{\textbf{EMScore} \mbox{\cite{shi2021emscore}}} & \mbox{Token} & \mbox{None} & \mbox{Off-the-shelf} & \yesmark &  & \mbox{Vid. Caption} \\
\mbox{\textbf{mPLUG-Owl-Video} \mbox{\cite{bansal2024videocon}}} & \mbox{Response} & \mbox{EEH, EQH, AVH, SRH, APH, TRH} & \mbox{Trained Model} & \yesmark &  & \mbox{Vid. Caption, Vid. QA} \\
\mbox{\textbf{FIFA} \mbox{\cite{jing2025fifa}}} & \mbox{Atomic Fact} & \mbox{EEH, AVH, SRH, APH} & \mbox{Off-the-shelf} &  & \yesmark & \mbox{V2T, T2V} \\
\mbox{\textbf{MTLA} \mbox{\cite{shalam2026propose}}} & \mbox{Temp. Window} & \mbox{None} & \mbox{Mechanistic Detection} & \yesmark &  & \mbox{Img Obj. Detection, Vid. Ground.} \\
\bottomrule
\end{tabular}%
}
\caption{Comparison of existing hallucination detection methods. \textit{Mechanistic Detection} uses internal model signals. \textit{Score} denotes continuous confidence and \textit{Label} denotes a discrete decision.}
\label{tab:detection_methods}
\end{table*}

\begin{table*}[!t]
\centering
\renewcommand{\arraystretch}{1.00}
\setlength{\tabcolsep}{0.7pt}
\fontsize{7.5}{8.6}\selectfont
\setlength{\extrarowheight}{0pt}
\renewcommand{\tabularxcolumn}[1]{m{#1}}
\begin{tabularx}{\textwidth}{@{}>{\raggedright\arraybackslash}X >{\centering\arraybackslash}m{47pt} >{\centering\arraybackslash}m{39pt} *{8}{>{\centering\arraybackslash}m{17pt}} >{\centering\arraybackslash}m{60pt} >{\centering\arraybackslash}m{52pt} >{\centering\arraybackslash}m{39pt}@{}}
\toprule
\multirow[c]{3}{=}[-5.5pt]{\centering\textbf{\mbox{Benchmark}}} &
\multirow[c]{3}{=}[-5.5pt]{\centering\textbf{\mbox{Detection Eval}}} &
\multirow[c]{3}{=}[-5.5pt]{\centering\textbf{\mbox{Granularity}}} &
\multicolumn{8}{c}{\textbf{\mbox{Type Annotation}}} &
\multirow[c]{3}{=}[-5.5pt]{\centering\textbf{\mbox{VLM Tasks}}} &
\multirow[c]{3}{=}[-5.5pt]{\centering\textbf{\mbox{\# Ques. / \# Vids.}}} &
\multirow[c]{3}{=}[-5.5pt]{\centering\textbf{\mbox{Adversarial}}} \\
\cmidrule(lr){4-11}
& & & \multicolumn{5}{c}{\textbf{\mbox{Ontology}}} & \multicolumn{3}{c}{\textbf{\mbox{Dynamic}}} & & & \\
\cmidrule(lr){4-8}\cmidrule(lr){9-11}
& & & \textbf{\mbox{EEH}} & \textbf{\mbox{ECH}} & \textbf{\mbox{EQH}} & \textbf{\mbox{AVH}} & \textbf{\mbox{SRH}} & \textbf{\mbox{APH}} & \textbf{\mbox{TRH}} & \textbf{\mbox{CPH}} & & & \\
\midrule
\textbf{VideoHallucer}\newline\mbox{\cite{wang2024videohallucerevaluatingintrinsicextrinsic}} &
$\times$ & Response & $\checkmark$ & $\times$ & $\times$ & $\checkmark$ & $\times$ & $\checkmark$ & $\checkmark$ & $\times$ &
Video QA & 1,800 / 948 & $\checkmark$ \\
\textbf{VidHalluc}\newline\mbox{\cite{li2025vidhalluc}} &
$\times$ & Response & $\times$ & $\times$ & $\times$ & $\times$ & $\times$ & $\checkmark$ & $\checkmark$ & $\checkmark$ &
Video QA & 9,295 / 5,002 & $\checkmark$ \\
\textbf{Vript-HAL}\newline\mbox{\cite{yang2024vript}} &
$\times$ & Response & $\checkmark$ & $\times$ & $\times$ & $\times$ & $\times$ & $\checkmark$ & $\times$ & $\times$ &
Caption & 122 / 122 & $\times$ \\
\textbf{EventHallusion}\newline\mbox{\cite{zhang2025eventhallusiondiagnosingeventhallucinations}} &
$\times$ & Response & $\times$ & $\times$ & $\times$ & $\times$ & $\times$ & $\checkmark$ & $\times$ & $\times$ &
Video QA & -- / 400 & $\times$ \\
\textbf{VideoHallu}\newline\mbox{\cite{li2025videohallu}} &
$\times$ & Response & $\times$ & $\times$ & $\checkmark$ & $\checkmark$ & $\checkmark$ & $\checkmark$ & $\checkmark$ & $\checkmark$ &
Video QA & 3,233 / 987 & $\checkmark$ \\
\textbf{ELV-Halluc}\newline\mbox{\cite{lu2025elv}} &
$\times$ & Response & $\times$ & $\times$ & $\times$ & $\checkmark$ & $\checkmark$ & $\checkmark$ & $\times$ & $\times$ &
Video QA & 4,800 / 200 & $\checkmark$ \\
\textbf{Dr.V-Bench}\newline\mbox{\cite{luo2025drvhierarchicalperceptiontemporalcognitionframework}} &
$\times$ & Response & $\checkmark$ & $\times$ & $\checkmark$ & $\checkmark$ & $\checkmark$ & $\checkmark$ & $\checkmark$ & $\times$ &
\mbox{Video QA, Caption} & 10,000 / 4,974 & $\checkmark$ \\
\textbf{OmniVCHall}\newline\mbox{\cite{xing2026omnivchall}} &
$\times$ & Response & $\checkmark$ & $\times$ & $\times$ & $\checkmark$ & $\checkmark$ & $\checkmark$ & $\checkmark$ & $\checkmark$ &
Video QA & 9,027 / 823 & $\checkmark$ \\
\midrule
\textbf{\VidHalLoc{} (ours)} &
$\checkmark$ & Response & $\checkmark$ & $\checkmark$ & $\checkmark$ & $\checkmark$ & $\checkmark$ & $\checkmark$ & $\checkmark$ & $\checkmark$ &
\mbox{Video QA, Caption} & 2,000 / 1,090 & $\checkmark$ \\
\bottomrule
\end{tabularx}
\caption{Comparison of \VidHalLoc{} with existing video hallucination benchmarks. \textit{Detection Eval} denotes whether detection methods are evaluated.}
\label{tab:detection_benchmarks}
\end{table*}

Our main contributions are summarized as follows:
\begin{itemize}[leftmargin=*,topsep=3pt,itemsep=3pt,parsep=0pt,partopsep=0pt]
    \item \textbf{A Diagnostic Framework} --- We organize video hallucinations into five ontological and three dynamic categories to evaluate detection reliability.

    \item \textbf{\VidHalLoc{}} --- We introduce a benchmark comprising 2,000 adversarial samples spanning Video Question Answering and Video Captioning, with a unified protocol for evaluating hallucination detectors.

    \item \textbf{\VideoHALO{}} --- We develop a multi-agent workflow for video benchmark construction, informed by harness engineering and supported by a unified communication protocol and hierarchical memory.
\end{itemize}

\section{Related Work}\label{sec:related_work}
\begingroup
\setlength{\parskip}{0pt}

\paragraph{From Image to Video Hallucination Detection}\equaldash{\textemdash}Image-oriented hallucination detection assesses either complete responses or finer semantic units~\cite{rohrbach2018object,petryk2024aloha,gunjal2024detecting,Park_2025_CVPR}. At the response level, GAVIE uses GPT-4 to evaluate an answer against textual descriptions of the image~\cite{liu2023mitigating}, and PAC-S measures overall image--text compatibility in an embedding space~\cite{sarto2023positive}. Fine-grained approaches instead examine individual claims within a response. FaithScore extracts atomic facts and verifies them against the image~\cite{jing2024faithscore}, and UNIHD checks individual claims using evidence from auxiliary tools~\cite{chen2024unified}.

Video detection must connect textual claims with evidence across frames and events. Embedding-based methods include PAC-S for image and video caption evaluation~\cite{sarto2023positive} and EMScore, which combines video--sentence matching with frame--word alignment~\cite{shi2021emscore}. For learned entailment, Owl-Con is the mPLUG-Owl-Video model fine-tuned on VideoCon to determine whether a video supports a textual claim~\cite{bansal2024videocon}. FIFA performs structured verification by extracting facts, modeling their dependencies, and checking them against video evidence~\cite{jing2025fifa}. Another line of work uses internal model signals: MTLA estimates grounding confidence by aggregating prediction-token attention within a proposed spatiotemporal region~\cite{shalam2026propose}.

These methods differ in their evaluation units, detection mechanisms, and task-specific protocols, making their reliability across hallucination types difficult to compare. \VidHalLoc{} addresses this gap by evaluating representative detection methods under a unified hallucination taxonomy and an adversarial evaluation protocol. Table~\ref{tab:detection_methods} provides the detailed method comparison.

\vspace{-1pt}\paragraph{Hallucination Benchmarks}\equaldash{\textemdash}\looseness=2 Hallucination benchmarks were initially developed to characterize when multimodal models produce unsupported content. In the image setting, POPE isolates object-existence errors, AMBER broadens the analysis to attributes and relations, and HallusionBench examines failures induced by visual illusions and misleading language contexts~\cite{li2023evaluating,wang2024amberllmfreemultidimensionalbenchmark,guan2024hallusionbench}. Video benchmarks introduce a different challenge because the relevant evidence may be distributed across frames and events. VidHalluc focuses on temporal ordering and event consistency, exposing errors that cannot be diagnosed from isolated frames~\cite{li2025vidhalluc}. ELV-Halluc moves the evaluation to long-form videos, where a claim may appear locally plausible while conflicting with evidence aggregated across distant segments~\cite{lu2025elv}. Dr.V-Bench instead provides fine-grained spatiotemporal grounding, allowing a hallucinated claim to be associated with the relevant interval and visual region~\cite{luo2025drvhierarchicalperceptiontemporalcognitionframework}. Table~\ref{tab:detection_benchmarks} compares the task categories, hallucination taxonomies, evaluation objectives, sample sizes, and adversarial characteristics across existing benchmarks. While existing benchmarks evaluate LVLM hallucinations within task-specific settings, they cannot ensure the reliability of detection techniques across diverse hallucination types. Furthermore, manually constructing video datasets becomes costly when fine-grained control over type-specific annotations is required~\cite{li2025vidhalluc,lu2025elv,luo2025drvhierarchicalperceptiontemporalcognitionframework}.

\par\endgroup
\vspace{-1pt}\section{\VidHalLoc{} Benchmark}\label{sec:vidhalloc_benchmark}
\looseness=-1 \vspace{-1pt}We present the \VidHalLoc{} benchmark of 2,000 instances to evaluate hallucination detectors, featuring coupled annotations of hallucination types across the core tasks of Video Question Answering (Video QA) and Video Captioning.

\subsection{Video Hallucination Types}\label{sec:hallucination_types}
Building upon prior observations of video misalignments~\cite{bai2024hallucination,chen2024unified,wang2024videohallucerevaluatingintrinsicextrinsic,li2025vidhalluc}, we present a comprehensive framework to facilitate a rigorous assessment of detection reliability. Specifically, we organize general video hallucinations into two top-level categories: Ontology and Dynamic. Ontology hallucination describes entity-related misalignments, encompassing objects, scenes, attributes, categories, spatial relations, and quantities. In contrast, Dynamic hallucination characterizes motion and temporal inconsistencies, specifically including actions, temporal order, and camera transitions.

\paragraph{Ontology Hallucination (OH)}\equaldash{\textemdash}Ontology hallucination describes entity-related misalignments across five core aspects: Entity Existence Hallucination (EEH): fabricates or explicitly denies the presence of an object or scene. Entity Category Hallucination (ECH): misidentifies the semantic category of a grounded entity (\textit{e.g., describing a basketball as a soccer ball}). Entity Quantity Hallucination (EQH): miscounts visible entities within a given interval. Attribute Value Hallucination (AVH): distorts an observable property of a grounded entity (\textit{e.g., describing a closed door as open, or a blue pen as red}). Static Relation Hallucination (SRH): misrepresents the spatial relationship between grounded entities.

\paragraph{Dynamic Hallucination (DH)}\equaldash{\textemdash}Dynamic hallucination describes motion and temporal inconsistencies across actions, temporal order, and camera transitions: Action Predicate Hallucination (APH): mischaracterizes the action or behavior of a grounded entity. Temporal Relation Hallucination (TRH): reverses or distorts the chronological order between valid events (\textit{e.g., event A occurs before event B, but the output reverses their order}). Camera Predicate Hallucination (CPH): misidentifies camera motions or editing operations (\textit{e.g., the camera zooms in while the video actually zooms out}).

\begin{figure*}[!t]
\centering
\makebox[\textwidth][c]{\includegraphics[width=0.92\textwidth]{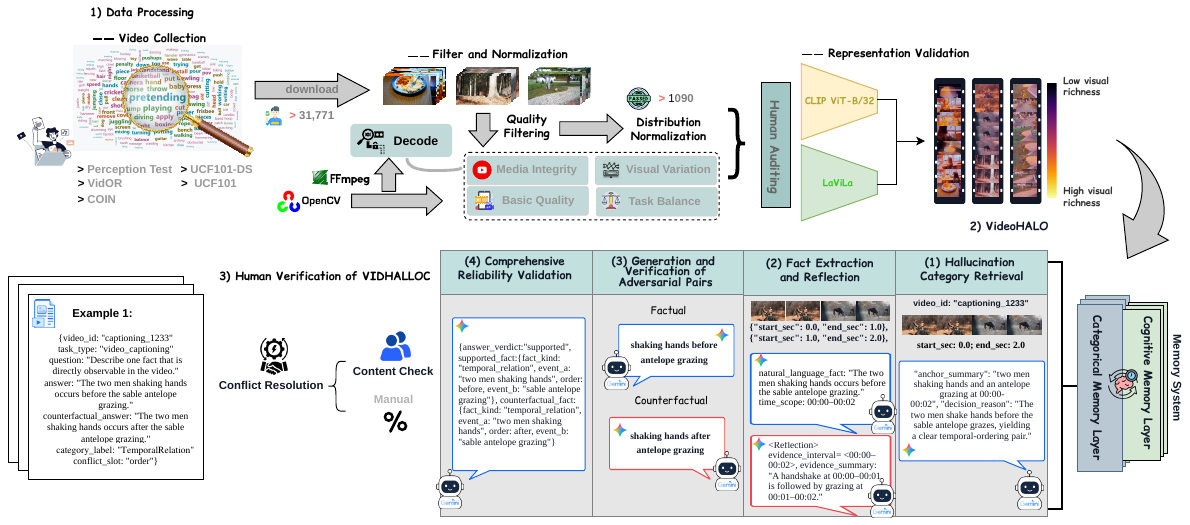}}
\caption{End-to-End Benchmark Construction\equaldash{---}The data processing pipeline curates 1,090 high-quality samples from an initial pool of 31,771 videos through rigorous decoding, filtering, distribution normalization, and human verification. To guarantee representativeness, CLIP ViT-B/32~\cite{radford2021clip} ensures dataset-wide visual diversity, while LaViLa~\cite{zhao2023lavila} captures temporal scene transitions within individual videos. Flowing from right to left, \VideoHALO{} orchestrates four collaborative stages: Hallucination Category Retrieval, Fact Extraction and Reflection, Generation and Verification of Adversarial Pairs, and Comprehensive Reliability Validation. Throughout this workflow, a dual-layer memory system provides a globally consistent cognitive foundation for all agents, while stage-specific records dynamically govern the information shared between roles. Finally, human reviewers independently audit a subset of the generated samples (Appendix~\ref{app:human_verification}).}
\label{fig:methodology_workflow}
\end{figure*}

\subsection{Data Processing}\label{sec:data_processing}
\paragraph{Video Collection}\equaldash{\textemdash}To provide a rich and diverse foundation for dataset construction, we curate a candidate pool of 31,771 real-world videos sourced from VidOR, COIN, Perception Test, UCF101-DS, and UCF101~\cite{vidorDataset2019,coinDataset2019,perceptionTest2023,ucf101ds2023,ucf101Dataset2012}. These sources encompass a wide spectrum of authentic visual content, covering entity relations, multistep instructional activities, general perception scenarios, and temporally bounded human actions.

\paragraph{Filter and Normalization}\equaldash{\textemdash}We process the initial video pool in three steps using FFmpeg~\cite{tomar2006ffmpeg} and OpenCV~\cite{opencv_library}. (1) Quality Filtering: we remove defective data, including files that are corrupted, duplicated, or unable to pass basic visual checks. (2) Distribution Normalization: we balance the dataset, preventing the over-representation of either static shots or dynamic transitions while maintaining a proper proportion of video sources and tasks. (3) Human Auditing: human reviewers check a random subset to confirm the automated decisions, yielding a refined pool of 1,090 videos ready for representation validation.

\paragraph{Representation Validation}\equaldash{\textemdash}Before feeding the candidate pool into \VideoHALO{}, we extract and validate the feature representations of every video to ensure two key qualities: wide visual diversity across the entire dataset and sufficient temporal changes within individual clips. Specifically, we use CLIP~\cite{radford2021clip} to extract global visual representations that examine overall semantic coverage~\cite{mcinnes2018umap}, and LaViLa~\cite{zhao2023lavila} to capture sequential temporal representations that measure scene variations over time. Figures~\ref{fig:data_processing_scatter} and~\ref{fig:data_processing_histograms} in Appendix~\ref{app:data_processing} report these checks to confirm the breadth and richness of the dataset.

\subsubsection{\VideoHALO{}}\label{sec:videohalo}
Inspired by Harness Engineering~\cite{zhong2026aiharnessengineering}, \VideoHALO{} automates video dataset construction by decomposing the annotation process into four executable sub-tasks. By equipping role-specific agents with a hierarchical memory system and a unified communication protocol~\cite{hong2024metagpt,wu2024autogen,qian2024chatdev,sumers2024coala}, our workflow simplifies complex benchmark engineering while strictly preserving quality. Figure~\ref{fig:methodology_workflow} summarizes this pipeline, with complete interfaces and rejection rules detailed in Appendix~\ref{app:workflow_implementation} (Section~\ref{app:multi_agent_workflow}).

\paragraph{Memory System}\equaldash{\textemdash}To ensure robust synchronization across the multi-agent workflow, \VideoHALO{} implements a hierarchical memory system comprising two foundational layers invoked during each agent call. (1) Systematic Cognitive Layer: The cognitive layer establishes overarching data construction boundaries and the exact protocols for synthesizing adversarial samples. (2) Categorical Memory Layer: The categorical layer supplies precise specifications for all hallucination types, indicating their conceptual boundaries, illustrative examples, and retrieval rules. Ultimately, this dual-layer architecture instills a universally consistent cognitive foundation, seamlessly guiding the entire pipeline to generate fine-grained, high-quality data (Appendix~\ref{app:memory_system}).

\paragraph{Coordinated Video Understanding Sub-tasks}\equaldash{\textemdash}\kern-0.025em\VideoHALO{} progresses through four collaborative stages. (1) Hallucination Category Retrieval: The planner agent scans the video to identify promising scenes. (2) Fact Extraction and Reflection: The extraction and reflection agents isolate and verify specific visual details. (3) Generation and Verification of Adversarial Pairs: Text-only agents utilize predefined templates (Table~\ref{tab:question_templates}) to formulate a targeted question and modify a single key detail to form a counterfactual statement, followed by a rigorous cross-check to ensure the broader context remains unaltered. (4) Comprehensive Reliability Validation: The monitor agent re-engages the visual modality to check the samples against video evidence. Appendix~\ref{app:coordinated_sub-tasks} details these task contracts.

\paragraph{Communication Protocol}\equaldash{\textemdash}\kern-0.025em\VideoHALO{} orchestrates multi-agent collaboration via a structured communication protocol. Rather than relying on open-ended dialogue, agents exchange outputs through standardized, schema-driven forms. To preserve task state, crucial fields validated in earlier stages are locked as permanent contextual states. Downstream agents evaluate this propagated information while being restricted from overwriting prior conclusions, ensuring consistent reasoning across the pipeline (detailed in Appendix~\ref{app:communication_protocol}).

\begin{table*}[!t]
    \centering
    
    \small
    \textbf{(a) Category Annotation Quality}\par\smallskip
    {\setlength{\tabcolsep}{4.2pt}
    \begin{tabular}{lrrrrrrrrr}
        \toprule
        \multirow{2}{*}{\mbox{Metric}} & \multicolumn{5}{c}{\textbf{\mbox{Ontology Hallucination}}} & \multicolumn{3}{c}{\textbf{\mbox{Dynamic Hallucination}}} & \multirow{2}{*}{\mbox{All}} \\\cmidrule(lr){2-6}\cmidrule(l{2pt}r{0pt}){7-9}& \mbox{EEH} & \mbox{ECH} & \mbox{EQH} & \mbox{AVH} & \mbox{SRH} & \mbox{APH} & \mbox{TRH} & \mbox{CPH} & \\
        \midrule
        Sample accuracy (\%) & 99.00 & 93.00 & 100.00 & 100.00 & 100.00 & 98.00 & 100.00 & 100.00 & 98.75 \\
        Worker agreement (\%) & 97.00 & 96.00 & 100.00 & 98.00 & 99.00 & 96.00 & 93.00 & 94.00 & 96.63 \\
        Cohen's $\kappa$ & 0.983 & 0.975 & 1.000 & 0.963 & 0.989 & 0.976 & 0.958 & 0.964 & 0.962 \\
        \bottomrule
    \end{tabular}}

    \vspace{0.65em}
    \textbf{(b) Construction Efficiency}\par\smallskip
    {\setlength{\tabcolsep}{9pt}
    \begin{tabular}{lcc}
        \toprule
        \mbox{Method} & \mbox{Throughput $\uparrow$} & \mbox{Unit price $\downarrow$} \\
        \midrule
        \mbox{Human workers} & 24.50 & 0.630 \\
        \VideoHALO{} & \textbf{41.45} & \textbf{0.194} \\
        \bottomrule
    \end{tabular}}
\caption{Data Quality and Construction Efficiency\equaldash{---}(a) Sample accuracy compares original benchmark categories with one author's independent reference labels. Agreement and Cohen's $\kappa$ compare the two workers. Per-category $\kappa$ uses one-vs-rest coding, whereas All reports multiclass $\kappa$ (Appendix~\ref{app:human_verification}). (b) Throughput measures accepted samples per hour, and unit price is reported in AUD per sample, covering both \VideoHALO{} API inference and human labor.}
    \label{tab:vidhalloc_verification}
    \label{tab:vidhalloc_efficiency}
\end{table*}

\subsubsection{External Human Audit}\label{sec:human_audit}

\looseness=-1 The human audit yielded a sample accuracy of 98.75\% against one author's independently assigned reference categories. The two workers achieved 96.63\% category agreement, with an overall multiclass Cohen's $\kappa$ of 0.962 (Table~\ref{tab:vidhalloc_verification}). Appendix~\ref{app:human_verification} details the sampling procedure, annotation protocol, metric definitions, and confidence intervals. Table~\ref{tab:vidhalloc_efficiency} also reports construction throughput and unit price, which, together with the audit results, indicate that \VideoHALO{} combines efficient benchmark construction with high data quality.

\subsection{Data Statistics}\label{sec:data_statistics}

\begin{table*}[!b]
\centering
\small
\setlength{\tabcolsep}{4.2pt}
\renewcommand{\arraystretch}{1.10}
\begin{tabular*}{\textwidth}{@{\extracolsep{\fill}}lcccccccc@{}}
\toprule
\multirow{2}{*}{\textbf{\mbox{Statistic}}} &
\multicolumn{5}{c}{\textbf{\mbox{Ontology}}} &
\multicolumn{3}{c}{\textbf{\mbox{Dynamic}}} \\
\cmidrule(lr){2-6}\cmidrule(l{0pt}r{0pt}){7-9}
& \textbf{\mbox{EEH}} & \textbf{\mbox{ECH}} & \textbf{\mbox{EQH}} & \textbf{\mbox{AVH}} & \textbf{\mbox{SRH}}
& \textbf{\mbox{APH}} & \textbf{\mbox{TRH}} & \textbf{\mbox{CPH}} \\
\midrule
\# Questions & 250 & 250 & 250 & 250 & 250 & 250 & 250 & 250 \\
\# Videos & 250 & 250 & 250 & 250 & 250 & 250 & 250 & 250 \\
Avg. question length (words) & 13.28 & 13.47 & 12.78 & 12.97 & 12.06 & 12.14 & 15.30 & 13.46 \\
Avg. video length (s) & 50.88 & 48.88 & 49.25 & 50.03 & 48.95 & 48.63 & 53.44 & 61.05 \\
\bottomrule
\end{tabular*}
\caption{Category-Level Statistics of \VidHalLoc{}\kern-0.025em\equaldash{---}The Ontology group comprises EEH, ECH, EQH, AVH, and SRH. The Dynamic group contains APH, TRH, and CPH. Every individual category includes exactly 250 adversarial hallucination samples. The reported video counts reflect category-specific assignments derived from a total of 1,090 unique source videos. Length statistics quantify the word counts of the textual queries and the temporal durations of their corresponding videos.}
\label{tab:vidhalloc_category_statistics}
\end{table*}

\VidHalLoc{} comprises 2,000 adversarial hallucination samples from 1,090 unique source videos, with exactly 250 instances in each category (Table~\ref{tab:vidhalloc_category_statistics}). The table reports category-level question and video counts, average question lengths, and average video durations. A source video can contribute samples to multiple categories. Average video durations range from 48.63 seconds for APH to 61.05 seconds for CPH; AVH and EEH both exceed 50 seconds, while TRH averages 53.44 seconds.

Figure~\ref{fig:vidhalloc_data_statistics} (left) in Appendix~\ref{app:data_processing} shows the numbers of Video QA and Captioning samples within each category. ECH has an equal split of 125 Video QA and 125 Captioning instances. The corresponding counts are 120 and 130 for AVH, and 145 and 105 for EEH. CPH has the largest difference between the two task counts, with 146 Video QA and 104 Captioning instances.

Average question lengths in Table~\ref{tab:vidhalloc_category_statistics} range from 12.06 words for SRH to 15.30 words for TRH, with EEH and ECH averaging 13.28 and 13.47 words, respectively. The word cloud in Figure~\ref{fig:vidhalloc_data_statistics} (right) displays vocabulary used in the benchmark questions, including task words such as summarize, identify, state, and describe.

\renewcommand{\dbltopfraction}{0.90}
\setcounter{dblbotnumber}{2}
\subsection{Evaluation}\label{sec:evaluation}
We assess the extent of alignment through three performance measures. These comprise Factual, Counterfactual, and Overall. For any given instance $i$, let the tuple $(d_i^F,d_i^C)\in\{\mathsf{S},\mathsf{R}\}^2$ denote the method decisions. The variable $d_i^F$ dictates whether the method supports ($\mathsf{S}$) or rejects ($\mathsf{R}$) the factual answer. The variable $d_i^C$ indicates the corresponding decision for the counterfactual answer. These joint decisions establish four exclusive outcome states:
{\setlength{\abovedisplayskip}{4pt}\setlength{\belowdisplayskip}{4pt}\setlength{\abovedisplayshortskip}{4pt}\setlength{\belowdisplayshortskip}{4pt}
\[
\begin{array}{c|cc}
 & d_i^C=\mathsf{R} & d_i^C=\mathsf{S} \\[2pt]
\hline
\noalign{\vskip 2pt}
d_i^F=\mathsf{S} & \text{Both correct }(o_i=1) & \text{Factual only} \\
d_i^F=\mathsf{R} & \text{Counterfactual only} & \text{Neither}
\end{array}
\]
}
The Factual accuracy quantifies the proportion of instances satisfying $d_i^F=\mathsf{S}$. The Counterfactual accuracy measures the proportion of instances satisfying $d_i^C=\mathsf{R}$. The Overall accuracy isolates the joint success rate. The criterion demands the "Both correct" state. The ideal outcome corresponds to the indicator function $o_i=\mathbf{1}\{d_i^F=\mathsf{S}\land d_i^C=\mathsf{R}\}$. This strict evaluation dictates simultaneous comprehension across both statements.

\vspace{-1pt}\section{Experiments}\label{sec:experiments}
\begin{table*}[!b]
\centering
\fontsize{8}{9}\selectfont
\renewcommand{\arraystretch}{1.05}
\setlength{\tabcolsep}{1.5pt}
\setlength{\extrarowheight}{0pt}
\setlength{\arrayrulewidth}{0.4pt}
\renewcommand{\tabularxcolumn}[1]{m{#1}}
\begin{tabularx}{\textwidth}{@{}>{\raggedright\arraybackslash}X|>{\centering\arraybackslash}m{48pt}>{\centering\arraybackslash}m{72pt}>{\centering\arraybackslash}m{30pt}|>{\centering\arraybackslash}m{36pt}>{\centering\arraybackslash}m{63pt}>{\centering\arraybackslash}m{36pt}@{}}
\hline
\multicolumn{4}{c}{} & \multicolumn{3}{c}{\textbf{\mbox{Accuracy on \VidHalLoc{}}}} \\
\cmidrule(l{1.5pt}r{1.5pt}){5-7}
\textbf{\mbox{Methods}} & \textbf{\mbox{LLM Params}} & \textbf{\mbox{Encoder}} & \textbf{\mbox{Frames}} & \textbf{\mbox{Factual $\uparrow$}} & \textbf{\mbox{Counterfactual $\uparrow$}} & \textbf{\mbox{Overall $\uparrow$}} \\
\methodfamilyrow{Commercial VLMs}
Gemini-3-Flash \mbox{\cite{google2025gemini3flash}} & -- & -- & 1fps & \accell{30}{\textbf{98.25}} & \accell{21}{84.63} & \accelllast{21}{\textbf{83.63}} \\
GPT-5 \mbox{\cite{openai2025gpt5}} & -- & -- & 1fps & \accell{21}{83.63} & \accell{26}{\underline{92.75}} & \accelllast{18}{79.50} \\
\methodfamilyrow{Open Source VLMs}
LLaVA-NeXT-Video \mbox{\cite{li2024llavanextinterleave}} & 7B & CLIP ViT-L/14 & 32 & \accell{16}{72.38} & \accell{6}{25.63} & \accelllast{2}{2.13} \\
Qwen3-VL-Instruct \mbox{\cite{bai2025qwen3vl}} & 8B & Qwen3-VL-ViT & 32 & \accell{23}{87.88} & \accell{21}{83.38} & \accelllast{16}{73.00} \\
Gemma-4-it \mbox{\cite{gemmateam2026gemma4}} & 12B & Unified & 32 & \accell{11}{55.88} & \accell{28}{\textbf{94.88}} & \accelllast{10}{52.63} \\
InternVL3.5 \mbox{\cite{wang2025internvl35}} & 14.8B & InternViT-300M & 32 & \accell{25}{90.88} & \accell{16}{72.13} & \accelllast{13}{64.88} \\
Qwen3.6-27B \mbox{\cite{qwenteam2026qwen36}} & 27B & Qwen3.5-Vision-Encoder & 32 & \accell{29}{95.50} & \accell{21}{84.50} & \accelllast{19}{\underline{81.00}} \\
Qwen3-Omni-Instruct \mbox{\cite{xu2025qwen3omni}} & 30B-A3B & Qwen3-Omni ViT & 32 & \accell{30}{\underline{98.13}} & \accell{9}{48.00} & \accelllast{9}{46.75} \\
\methodfamilyrow{Video Agents}
VideoAgent \mbox{\cite{fan2024videoagentmemory}} & -- & -- & -- & \accell{7}{36.63} & \accell{24}{89.75} & \accelllast{6}{31.50} \\
VideoHV-Agent \mbox{\cite{wang2026think}} & -- & -- & 1fps & \accell{19}{80.75} & \accell{10}{53.13} & \accelllast{7}{39.75} \\
Deep Video Discovery \mbox{\cite{zhang2025deepvideodiscovery}} & -- & -- & 2fps & \accell{23}{88.38} & \accell{19}{81.00} & \accelllast{16}{71.50} \\
\methodfamilyrow{Detection Methods}
PAC-S \mbox{\cite{sarto2023positive}} & -- & CLIP ViT-B/32 & -- & \accell{9}{47.75} & \accell{11}{56.88} & \accelllast{2}{7.25} \\
EMScore \mbox{\cite{shi2021emscore}} & -- & CLIP ViT-B/32 & -- & \accell{14}{66.63} & \accell{7}{37.13} & \accelllast{2}{6.50} \\
Owl-Con (mPLUG-Owl-7B-Video) \mbox{\cite{bansal2024videocon}} & 7B & CLIP ViT-L/14 & 32 & \accell{14}{67.50} & \accell{12}{59.25} & \accelllast{6}{33.13} \\
FIFA \mbox{\cite{jing2025fifa}} & -- & -- & 1fps & \accell{8}{44.25} & \accell{24}{88.88} & \accelllast{7}{34.63} \\
\hline
\end{tabularx}
\caption{Performance Comparison of Existing Methods on \VidHalLoc{}\kern-0.025em\equaldash{---}The numbers in the table represent accuracy percentages (\%). \textbf{Bold} numbers denote the best performance, and \underline{underlined} numbers indicate the second-best performance. Appendix~\ref{app:main_experiment_settings} reports the input and execution settings.}
\label{tab:main_results}
\end{table*}

\begin{figure*}[!b]
\centering
\includegraphics[width=0.94\textwidth]{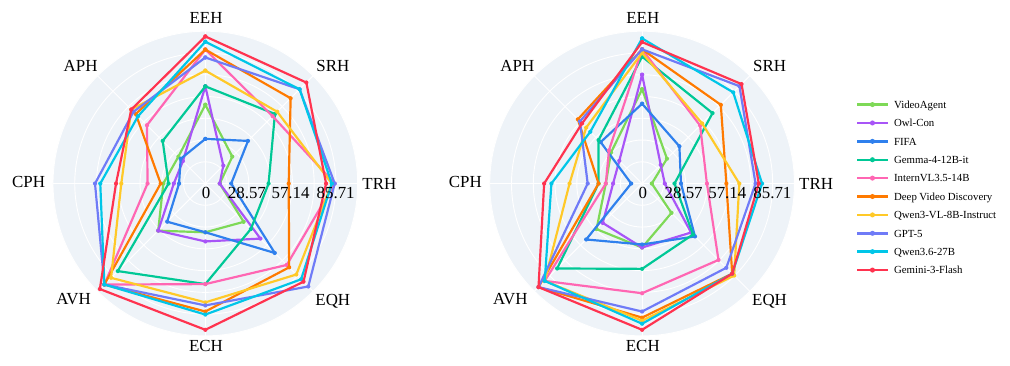}
\caption{Comparative Results on \VidHalLoc{} Across Various Hallucination Types\equaldash{---}Left: Video QA. Right: Caption evaluation. Each axis indicates the overall metric for a distinct hallucination type. TRH, APH, and CPH represent temporal sequence, action, and camera transition hallucinations, respectively, while EEH, ECH, EQH, AVH, and SRH correspond to entity-level hallucinations regarding objects, categories, quantities, attributes, and spatial relations.}
\label{fig:category_accuracy_radar}
\end{figure*}
\vspace{-1pt}We evaluate fifteen methods on \VidHalLoc{}, including two commercial models, six open-source models, three video agents, and four detection methods. Alongside the four dedicated detectors, we employ LVLMs and video agents as independent evaluators to determine whether each candidate response is supported by the video and correctly answers the question. Our results reveal that most LVLMs and video agents exhibit notable vulnerabilities on \VidHalLoc{} (Section~\ref{eval_LVLMs}). We then assess the performance of four representative detection methods on \VidHalLoc{}, showing the need to improve the reliability of these evaluators against various types of hallucinations (Section~\ref{eval_evaluators}). During inference, we preserve each model’s original configuration, including conversation mode, hyperparameters, and frame count. Following standard practices~\cite{cheng2024videollama,zhang2025videollama}, we set temperature to 0, top-\(k\) to 1, and disable stochastic sampling for all open-source models to avoid randomness in response generation. For commercial models, we sample frames at 1 fps. For video agents and detection methods, we use the original configuration from their paper. Implementation details and threshold settings are in Appendix~\ref{app:main_experiment_settings}. Appendix~\ref{app:statistical_analysis} reports the corresponding uncertainty analysis.
\subsection{Evaluation on \VidHalLoc{}}\label{eval_LVLMs}

\looseness=1 Table~\ref{tab:main_results} and Figure~\ref{fig:category_accuracy_radar} present the performance of all tested LVLMs and video agents on \mbox{\VidHalLoc{}}, showing accuracy as percentages. Across both Video QA and video captioning tasks, we observe that the Overall accuracy of most LVLMs and all three video agents score at least 17.60\% and 15.60\% lower on DH compared to OH, respectively (Appendix~\ref{app:method_level_intervals}, Table~\ref{tab:method_bootstrap}). This significant difference is due to the inherent focus of OH, which assesses stable factual elements, such as objects, categories, attributes, and spatial relations, that are easily verifiable within short video clips. In contrast, DH forces models to continuously reason across frames to differentiate detailed actions, event order, and camera transitions. Furthermore, due to the design of adversarial candidates, the factual and counterfactual statements become proximate in both visual and semantic spaces. Consequently, LVLMs and agents are more prone to \mbox{confusion and errors.}

We also observe that model scale does not correlate directly with performance. Commercial models generally outperform open-source models across most tasks, with Gemini-3-Flash~\cite{google2025gemini3flash} standing out in particular. However, Gemini-3-Flash still falls short on several DH tasks, scoring 55.77\% in APH Video Captioning and 58.62\% in CPH Video QA. In contrast, GPT-5~\cite{openai2025gpt5} does not consistently surpass the best open-source models. For instance, the Overall accuracy of 35.71\% in CPH Video Captioning is substantially lower than the 59.52\% achieved by Qwen3.6-27B~\cite{qwenteam2026qwen36}. These results highlight the need for improvement even in top-tier commercial models.

\renewcommand{\dbltopfraction}{0.70}
\subsection{Evaluation of Detection Methods on \VidHalLoc{}}\label{eval_evaluators}
To evaluate the effectiveness of detection methods in evaluating hallucinations, we use PAC-S~\cite{sarto2023positive}, EMScore~\cite{shi2021emscore}, Owl-Con~\cite{bansal2024videocon}, and FIFA~\cite{jing2025fifa} as baselines.
Under the evaluated protocol, none of the four dedicated detectors correctly resolves a majority of adversarial attacks, and performance deteriorates further on dynamic hallucinations (Appendix~\ref{app:threshold_settings}, Tables~\ref{tab:score_diagnostics}--\ref{tab:category_score_ohdh}).

EMScore~\cite{shi2021emscore} and PAC-S~\cite{sarto2023positive} achieve Overall scores of 16\% and 18\% respectively in ECH Video Captioning against a complete failure of 0\% in TRH Video Captioning. Although FIFA~\cite{jing2025fifa} and Owl-Con~\cite{bansal2024videocon} reach peak scores of 64.44\% in EQH Video QA and 71.43\% in EEH Video Captioning, respectively, this performance fails to generalize. Both methods experience severe accuracy drops in other tasks, with FIFA plunging to a mere 7.14\% in CPH Video Captioning and Owl-Con falling to 9.43\% in TRH Video QA.

\looseness=1 The evaluated methods assess video--text consistency through distinct mechanisms. Among embedding-based methods, PAC-S~\cite{sarto2023positive} computes visual--text similarity, and EMScore~\cite{shi2021emscore} combines global video--sentence matching with frame--word alignment. Neither explicitly encodes event order in this evaluation, which may contribute to their weak TRH results. Owl-Con is the mPLUG-Owl-Video entailment model trained with VideoCon contrastive captions~\cite{bansal2024videocon}, and its performance varies across hallucination categories. FIFA~\cite{jing2025fifa} decomposes responses into atomic facts before verifying them with multimodal evidence from off-the-shelf models. Because its output depends on both decomposition and evidence verification, errors at either stage can affect the final decision. Overall, these category-level variations indicate that embedding similarity, learned entailment, and structured verification transfer unevenly under the unified \VidHalLoc{} protocol (Appendices~\ref{app:threshold_settings} and~\ref{app:qualitative_examples}). 

\section{Conclusion}\label{limits}
\vspace{-2pt}We introduce \VidHalLoc{} to evaluate video hallucination detectors using adversarial candidates across various hallucination categories, and the \VideoHALO{} workflow for efficient benchmark construction through multiple agents. Mean Overall accuracy across fifteen evaluated systems is lower on dynamic hallucinations than on ontology types. The four dedicated detectors achieve a peak Overall accuracy of 34.63\%, trailing the top LVLM at 83.63\%. These findings establish dynamic hallucinations as a primary target for detector development and support an integrated evaluation of factual acceptance and counterfactual rejection.

\vspace{-1pt}\section*{Limitations}
VidHalLoc{} focuses on adversarial attacks differing in a single targeted detail. This design supports diagnosis by hallucination type but unconstrained model outputs may contain multiple interacting errors. The balanced category distribution may also diverge from frequencies encountered in practical deployments. Extending the benchmark with authentic responses would enable the evaluation of compound errors under target distributions.

The benchmark covers Video QA and Video Captioning using 1,090 videos from five public datasets with average durations per category ranging from 48.63 to 61.05 seconds. Coverage remains untested on longer recordings where relevant evidence spans distant segments. Extending the evaluation to these formats would test detector capacity to integrate distant evidence and verify event relations.

The evaluated systems employ varied visual encoders and frame sampling strategies alongside distinct language models and inference pipelines. The results characterize reliability under the reported configurations but fail to isolate the contribution of individual components. Complementary evaluations varying a single factor would help identify the specific design choices affecting detection reliability.

\bibliography{aaai2026}
\appendix
\flushbottom
\makeatletter
\setlength{\@fptop}{0pt}
\setlength{\@fpsep}{12pt plus 2pt minus 2pt}
\setlength{\@fpbot}{0pt plus 1fil}
\setlength{\@dblfptop}{0pt}
\setlength{\@dblfpsep}{12pt plus 2pt minus 2pt}
\setlength{\@dblfpbot}{0pt plus 1fil}
\makeatother
\setcounter{secnumdepth}{3}
\section*{Appendices}
To supplement the main text, the appendices document the \VideoHALO{} framework. Furthermore, we provide uncertainty analyses of the human audit and the main experimental results, alongside qualitative examples.

\section{Data Processing}\label{app:data_processing}
\paragraph{Video Collection}\equaldash{\textemdash}We collect 31,771 source videos from five public datasets, consisting entirely of authentic recordings and purpose-recorded scenes. Specifically, VidOR contributes YFCC100M footage featuring dense annotations of object trajectories, spatial relations, and interactions \cite{vidorDataset2019}. COIN provides YouTube instructional videos divided into temporally localized steps across 180 multistep tasks and 12 daily-life domains \cite{coinDataset2019}. The Perception Test dataset adds scripted real-world scenes recorded by around 100 global participants to probe memory, abstraction, physical reasoning, and semantic understanding \cite{perceptionTest2023}. UCF101-DS incorporates videos exhibiting naturally occurring shifts in actors, viewpoints, environments, occlusions, speed, and visual styles \cite{ucf101ds2023}. UCF101 supplies realistic YouTube footage spanning 101 classes of body motion, object interaction, interpersonal activity, musical performance, and sports \cite{ucf101Dataset2012}. Together, these sources combine spontaneous online footage with purpose-recorded scenarios, ensuring rich variation in scene composition, temporal organization, interaction complexity, and capture conditions.

\paragraph{Filter and Normalization}\equaldash{\textemdash}We detail the video processing pipeline, executed in three steps using FFmpeg~\cite{tomar2006ffmpeg} and OpenCV~\cite{opencv_library}. (1) For Quality Filtering: FFmpeg performs full decoding to eliminate duplicate videos and ensure media integrity, while OpenCV is utilized to guarantee the visual richness of the decoded content. (2) For Distribution Normalization: we quantify visual dynamics using consecutive pixel differences to assign motion scores. We then apply stratified sampling based on these scores to prevent dataset biases toward purely static shots or abrupt transitions, while controlling the composition of video sources and tasks. (3) Finally, in Human Auditing: We employ point estimation via random sampling, where independent reviewers evaluate a subset to infer the global quality threshold, ultimately yielding a refined pool of 1,090 videos ready for representation validation.

\begin{figure*}[tp]
\centering
\includegraphics[height=0.35\textwidth]{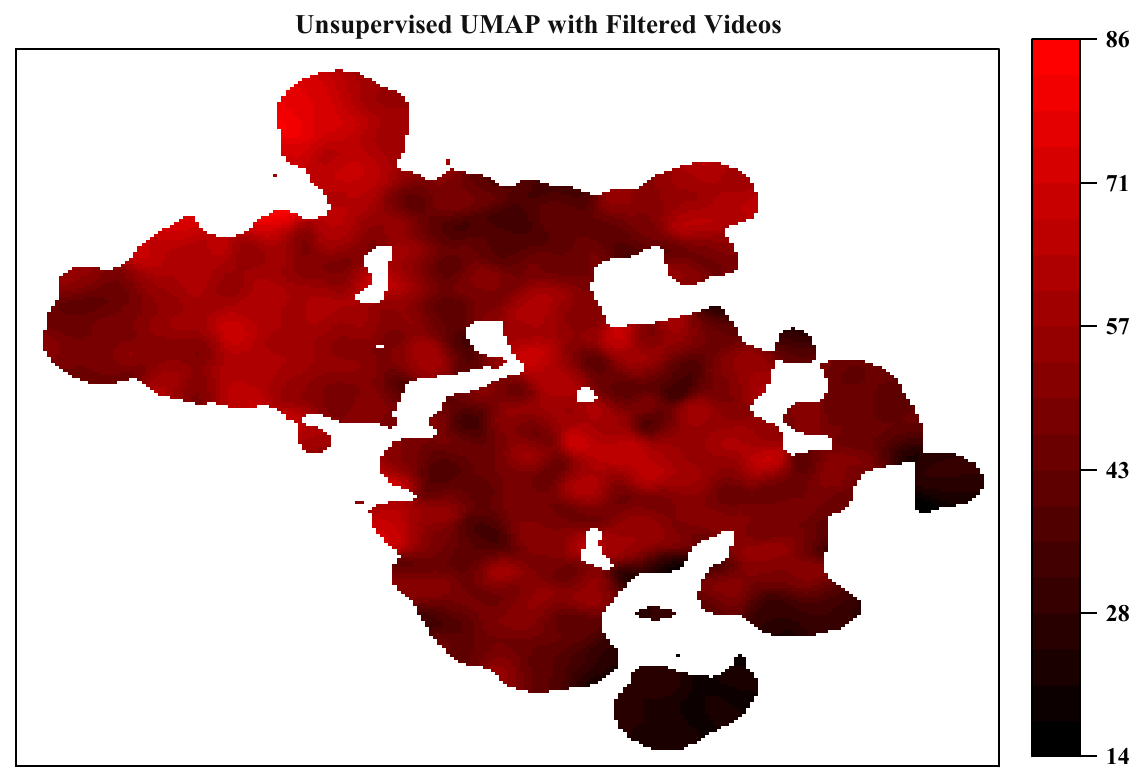}\hspace{0.008\textwidth}%
\includegraphics[height=0.35\textwidth]{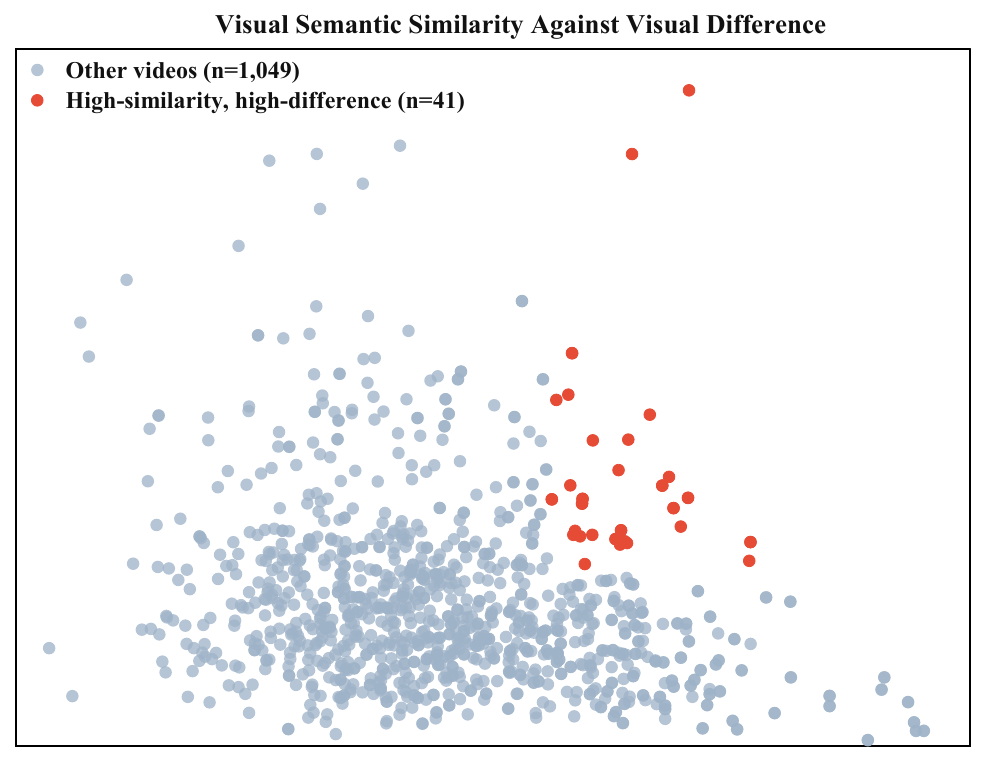}
\caption{Representations of Inter-Video Visual Diversity Using CLIP~\cite{radford2021clip}\equaldash{---}Left: global semantic coverage verifying dataset-wide visual richness. Right: joint analysis comparing inter-video nearest-neighbor similarity against within-video visual variations.}
\label{fig:data_processing_scatter}
\end{figure*}

\begin{figure*}[!t]
\centering
\includegraphics[height=0.29\textwidth]{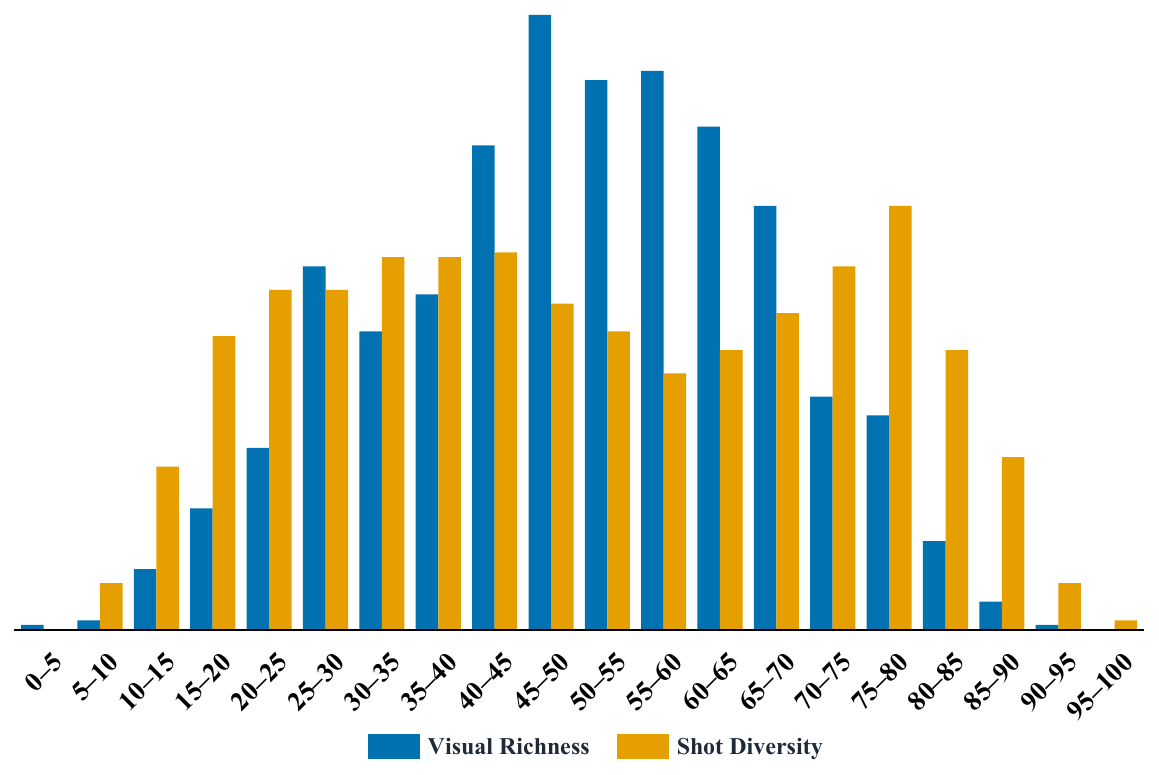}\hspace{0.008\textwidth}%
\includegraphics[height=0.29\textwidth]{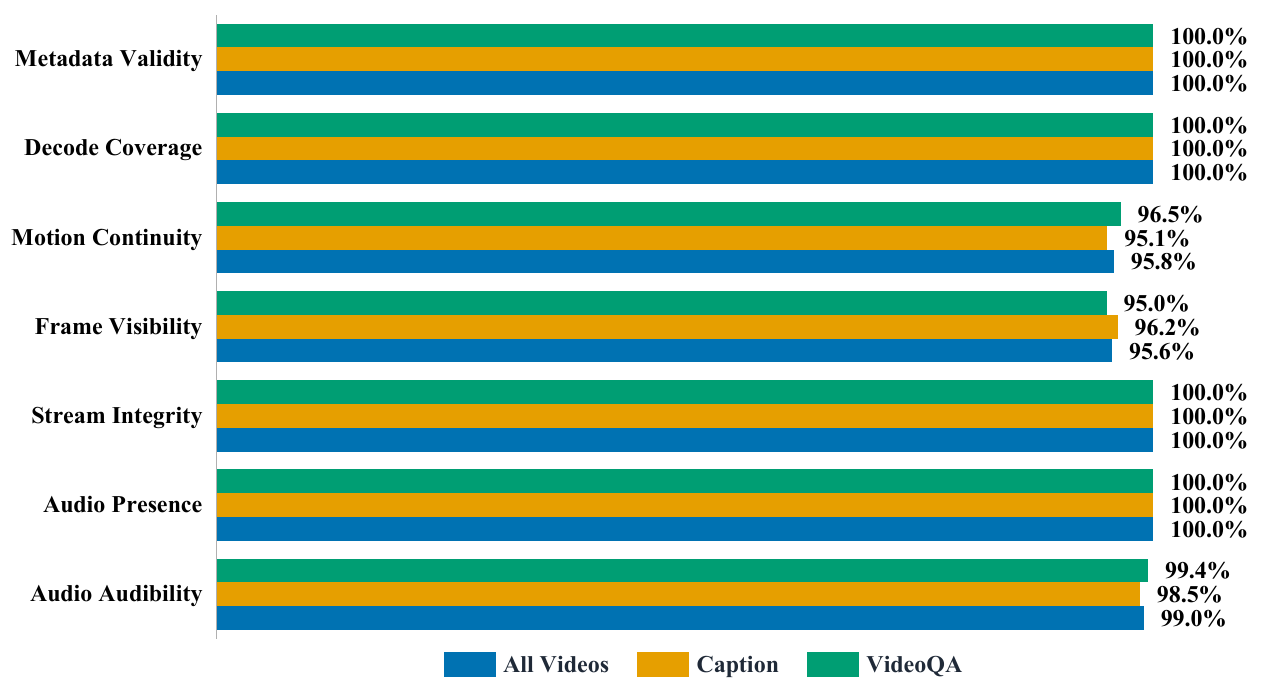}
\caption{Representations of Intra-Video Dynamics and Media Integrity\equaldash{---}Left: within-video visual richness evaluated via CLIP~\cite{radford2021clip}, contrasted with temporal shot diversity across sequential windows captured by LaViLa~\cite{zhao2023lavila}. Right: media integrity pass rates across data subsets.}
\label{fig:data_processing_histograms}
\end{figure*}

\paragraph{Representation Validation}\equaldash{\textemdash}To validate representations before \VideoHALO{}, we assess dataset-wide diversity and within-video temporal richness. (1) For inter-video breadth: CLIP ViT-B/32~\cite{radford2021clip} features from 8 uniformly sampled frames per video are mean-pooled into global vectors. Subsequent cosine similarity, nearest-neighbor, and UMAP~\cite{mcinnes2018umap} analyses confirm broad semantic coverage. (2) For intra-video dynamics: LaViLa~\cite{zhao2023lavila} encodes four frames across up to 12 temporal windows per clip. Evaluating adjacent and pairwise window distances alongside centroid dispersion verifies that each video exhibits genuine scene progression rather than static content. Appendix Figures~\ref{fig:data_processing_scatter} and~\ref{fig:data_processing_histograms} report these complementary checks.

\begin{figure*}[!t]
\centering
\includegraphics[height=0.25\textwidth]{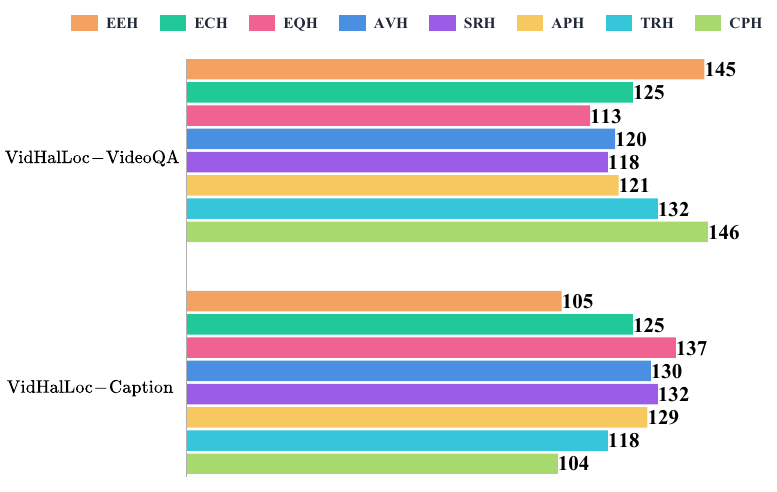}\hspace{0.008\textwidth}%
\includegraphics[height=0.25\textwidth]{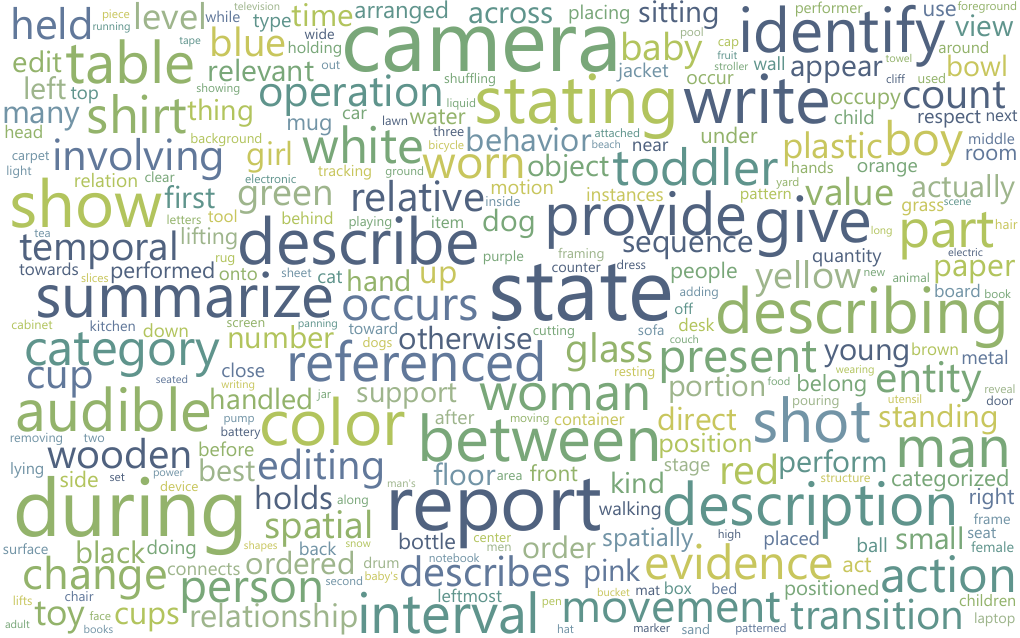}
\caption{Data Composition of \VidHalLoc{}\kern-0.025em\equaldash{---}Left: Number of data points per each hallucination type in the \VidHalLoc{}--VideoQA, \VidHalLoc{}--Caption. Right: word cloud of benchmark questions.}
\label{fig:vidhalloc_data_statistics}
\end{figure*}

\section{Multi-Agent Video Data Construction Workflow}\label{app:workflow_implementation}
Building upon the principles of Harness Engineering~\cite{zhong2026aiharnessengineering} and multi-agent orchestration~\cite{hong2024metagpt,wu2024autogen,qian2024chatdev}, \VideoHALO{} structures benchmark construction as a rigorous workflow centered around four verifiable sub-tasks. To execute this pipeline, we deploy a suite of specialized agents with strict operational boundaries. The \textit{Planner Agent} initially identifies category-specific opportunities. Subsequently, the \textit{Extraction} and \textit{Reflection Agents} collaborate to propose and independently validate atomic visual facts. Using these verified facts, the \textit{Generation} and \textit{Verification Agents} construct and back-parse coupled responses without direct video access. Finally, the \textit{Monitor Agent} executes a definitive video-grounded audit. By compartmentalizing discovery, generation, and verification, this architecture enforces strict internal checks and balances. It prevents systemic confirmation bias during data creation while preserving a fully traceable link from every finalized sample back to its visual evidence. The complete implementation is available in our \href{https://github.com/wesfggfd/VideoHALO-A-Human-Proxy-Agent-for-Fine-Grained-Video-Hallucination-Benchmarking-and-Evaluation}{project repository}.

\subsection{Memory System}\label{app:memory_system}
\begin{figure*}[!t]
\centering
\noindent\makebox[\linewidth][c]{%
\parbox{\linewidth}{%
\hrule
\vspace{0.14em}
{\raggedright\footnotesize\sffamily\bfseries SYSTEMATIC COGNITIVE MEMORY LAYER\par}
\vspace{0.14em}
\hrule}}\par
\vspace{0.04em}
\includegraphics[width=\linewidth]{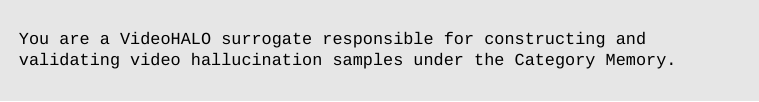}\par
\vspace{-0.3em}
\includegraphics[width=\linewidth]{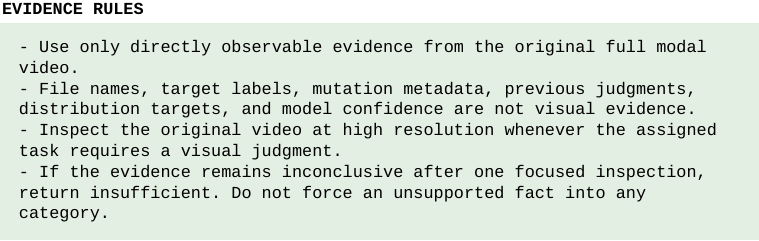}\par
\vspace{-0.3em}
\includegraphics[width=\linewidth]{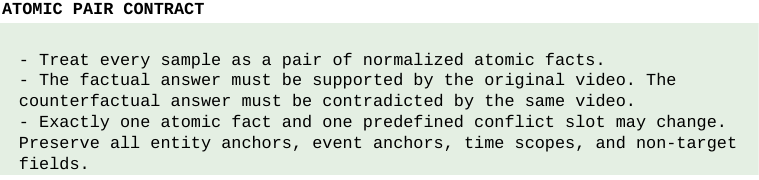}\par
\vspace{-0.3em}
\includegraphics[width=\linewidth]{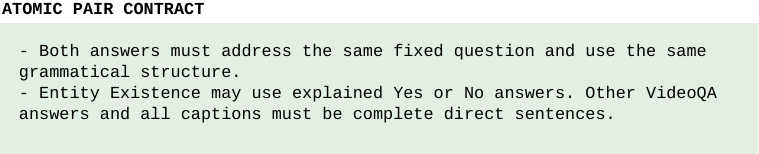}\par
\vspace{-0.3em}
\includegraphics[width=\linewidth]{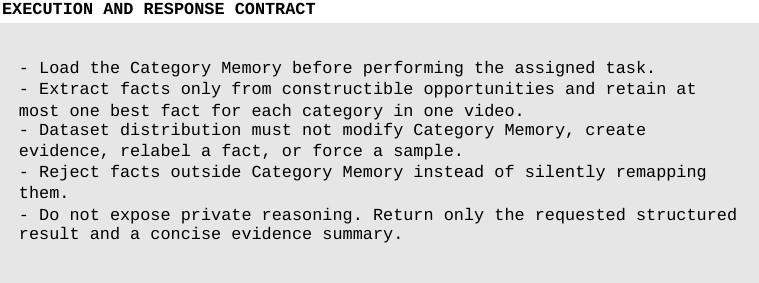}\par
\hrule height 0.4pt\relax
\caption{Systematic Cognitive Memory Used by \VideoHALO{}\kern-0.025em\equaldash{---}It standardizes the execution specifications and declares the data generation boundaries within multi-agent systems.}
\label{fig:cognitive_memory_1}
\end{figure*}

To ensure robust synchronization across the pipeline, \VideoHALO{} implements a hierarchical memory system~\cite{sumers2024coala} comprising two foundational layers invoked during each agent call. (1) Systematic Cognitive Memory Layer: The cognitive layer establishes overarching data construction boundaries and operational constraints throughout the orchestrated sub-tasks. (2) Categorical Memory Layer: The categorical layer provides precise specifications for various hallucination types, detailing their exact definitions, conceptual boundaries, illustrative examples, and retrieval rules. By aligning the multi-agent collaborative workflow in this dual-layer architecture, we ensure that all role-specific agents adhere to globally consistent annotation guidelines, thereby effectively prompting fine-grained data generation. Figures~\ref{fig:cognitive_memory_1}, \ref{fig:cognitive_memory}, \ref{fig:category_memory_1}, and~\ref{fig:category_memory_2} present the exact specifications for both layers.

\subsection{Coordinated Video Understanding Sub-tasks}\label{app:coordinated_sub-tasks}
As detailed in the corresponding figures, the \VideoHALO{} pipeline executes four interdependent sub-tasks. (1) Hallucination Category Retrieval (HCR): The planner agent leverages the dual-layer memory system to identify reasonable intervals for counterfactual fabrication, immediately discarding weak candidates (Figure~\ref{fig:hcr_specification}). (2) Fact Extraction and Reflection (FER): Building on these refined intervals, the extraction agent isolates a grounded atomic fact. The reflection agent subsequently corroborates this statement against the video, confirming both its temporal correctness and suitability for logical alteration (Figure~\ref{fig:fer_specification}). (3) Generation and Verification of Adversarial Pairs (GVP): Shifting to a text-only stage, the generation agent modifies a single key detail, utilizing predefined templates (Table~\ref{tab:question_templates}) to formulate the corresponding question and construct the conflicting statement. The verification agent subsequently back-parses adversarial responses into structured facts to validate the structural correctness of the samples (Figure~\ref{fig:gvp_specification}). (4) Comprehensive Reliability Validation (CRV): Closing the loop, the monitor agent re-engages the visual modality to rigorously verify the ultimate reliability of the generated samples. By conducting a final visual audit that allows at most one targeted re-inspection, this agent ensures the accepted data strictly maintains factual support and single-detail consistency (Figure~\ref{fig:crv_specification}).

\begingroup
\fontsize{8.4}{9.4}\selectfont
\setlength{\tabcolsep}{2.6pt}
\renewcommand{\arraystretch}{1.00}
\setlength{\LTleft}{0pt plus 1fill}
\setlength{\LTright}{0pt plus 1fill}
\setlength{\LTcapwidth}{\dimexpr0.97\textwidth+2\tabcolsep\relax}
\setlength{\LTpre}{2pt}

\begin{table*}[!t]
\centering
\small
\setlength{\tabcolsep}{2.6pt}
\renewcommand{\arraystretch}{1.06}

\begin{tabularx}{\textwidth}{@{}>{\centering\arraybackslash}p{0.07\textwidth}>{\raggedright\arraybackslash}X@{}}
\multicolumn{2}{c}{\textbf{Table~\ref{tab:question_templates}, Part I: \VidHalLoc{} Video QA categories EEH--SRH}} \\[2.5pt]
\toprule
\textbf{\mbox{Type}} & \textbf{\mbox{Exact source templates}} \\
\midrule
\multirow{6}{*}{\textbf{EEH}} & (1) \textbf{Is \textnormal{\texttt{\{entity\}}} present during the referenced part of the video?} \\*
 & (2) \textbf{Does the video show \textnormal{\texttt{\{entity\}}} in the referenced interval?} \\*
 & (3) \textbf{Can \textnormal{\texttt{\{entity\}}} be observed in this part of the video?} \\*
 & (4) \textbf{Is \textnormal{\texttt{\{entity\}}} visible or otherwise directly observable here?} \\*
 & (5) \textbf{Does \textnormal{\texttt{\{entity\}}} appear in the relevant portion of the video?} \\*
 & (6) \textbf{Is there direct video evidence that \textnormal{\texttt{\{entity\}}} is present?} \\
\cmidrule(lr){1-2}
\multirow{6}{*}{\textbf{ECH}} & (1) \textbf{What category does \textnormal{\texttt{\{entity\}}} belong to in the video?} \\*
 & (2) \textbf{What type of entity is \textnormal{\texttt{\{entity\}}}?} \\*
 & (3) \textbf{How should \textnormal{\texttt{\{entity\}}} be categorized based on the video?} \\*
 & (4) \textbf{Which entity category best describes \textnormal{\texttt{\{entity\}}}?} \\*
 & (5) \textbf{What kind of thing is \textnormal{\texttt{\{entity\}}} in this video?} \\*
 & (6) \textbf{Based on the video, what is the category of \textnormal{\texttt{\{entity\}}}?} \\
\cmidrule(lr){1-2}
\multirow{6}{*}{\textbf{EQH}} & (1) \textbf{How many \textnormal{\texttt{\{entity\_set\}}} are visible in the referenced interval?} \\*
 & (2) \textbf{What is the number of visible \textnormal{\texttt{\{entity\_set\}}} in this part of the video?} \\*
 & (3) \textbf{Count the \textnormal{\texttt{\{entity\_set\}}} shown in the referenced interval.} \\*
 & (4) \textbf{What quantity of \textnormal{\texttt{\{entity\_set\}}} can be observed here?} \\*
 & (5) \textbf{How many instances of \textnormal{\texttt{\{entity\_set\}}} does this interval show?} \\*
 & (6) \textbf{What count does the video support for \textnormal{\texttt{\{entity\_set\}}}?} \\
\cmidrule(lr){1-2}
\multirow{6}{*}{\textbf{AVH}} & (1) \textbf{What is the \textnormal{\texttt{\{attribute\_key\}}} of \textnormal{\texttt{\{entity\}}} in the video?} \\*
 & (2) \textbf{How would you describe the \textnormal{\texttt{\{attribute\_key\}}} of \textnormal{\texttt{\{entity\}}}?} \\*
 & (3) \textbf{Which \textnormal{\texttt{\{attribute\_key\}}} value is directly observable for \textnormal{\texttt{\{entity\}}}?} \\*
 & (4) \textbf{What \textnormal{\texttt{\{attribute\_key\}}} does the video show for \textnormal{\texttt{\{entity\}}}?} \\*
 & (5) \textbf{Which value describes the \textnormal{\texttt{\{attribute\_key\}}} of \textnormal{\texttt{\{entity\}}}?} \\*
 & (6) \textbf{Based on the visible evidence, what is the \textnormal{\texttt{\{attribute\_key\}}} of \textnormal{\texttt{\{entity\}}}?} \\
\cmidrule(lr){1-2}
\multirow{6}{*}{\textbf{SRH}} & (1) \textbf{Where is \textnormal{\texttt{\{subject\}}} relative to \textnormal{\texttt{\{object\}}}?} \\*
 & (2) \textbf{What is the spatial relation between \textnormal{\texttt{\{subject\}}} and \textnormal{\texttt{\{object\}}}?} \\*
 & (3) \textbf{How is \textnormal{\texttt{\{subject\}}} positioned with respect to \textnormal{\texttt{\{object\}}}?} \\*
 & (4) \textbf{What position does \textnormal{\texttt{\{subject\}}} occupy relative to \textnormal{\texttt{\{object\}}}?} \\*
 & (5) \textbf{How are \textnormal{\texttt{\{subject\}}} and \textnormal{\texttt{\{object\}}} spatially arranged?} \\*
 & (6) \textbf{Which spatial relationship holds between \textnormal{\texttt{\{subject\}}} and \textnormal{\texttt{\{object\}}}?} \\
\bottomrule
\end{tabularx}
\caption{Task template candidates used in \VidHalLoc{}\kern-0.025em\equaldash{---}Part I details the exact Video QA formats for EEH, ECH, EQH, AVH, and SRH. Following this, Part II presents APH, TRH, CPH, alongside eight caption formats drawn from a universal bank independent of specific categories. For APH specifically, the system employs two distinct banks of six templates: one tailored for facts conditioned on objects and another for actions lacking a direct object. Regarding notation, bold text represents fixed phrasing, whereas monospaced braces denote fields requiring instantiation. Ultimately, a deterministic key selects a single template before generating both the factual and counterfactual answers, guaranteeing that variations occur exclusively within the designated conflict slot.}\label{tab:question_templates}
\end{table*}
\endgroup

\begin{table*}[!t]
\begingroup
\fontsize{8.4}{9.4}\selectfont
\setlength{\tabcolsep}{2.6pt}
\renewcommand{\arraystretch}{1.00}
\setlength{\aboverulesep}{0.45ex}
\setlength{\belowrulesep}{0.45ex}
\setlength{\LTleft}{0pt plus 1fill}
\setlength{\LTright}{0pt plus 1fill}
\centering
\small
\setlength{\tabcolsep}{2.6pt}
\renewcommand{\arraystretch}{1.06}
\begin{tabularx}{\textwidth}{@{}>{\centering\arraybackslash}p{0.07\textwidth}>{\raggedright\arraybackslash}X@{}}
\multicolumn{2}{c}{\textbf{Table~\ref{tab:question_templates}, Part II: APH, TRH, CPH, and caption templates}} \\[2.5pt]
\toprule
\textbf{\mbox{Type}} & \textbf{\mbox{Exact source templates}} \\
\midrule
\multirow[c]{12}{*}{\textbf{APH}} & (1) \textbf{What action does \textnormal{\texttt{\{subject\}}} perform involving \textnormal{\texttt{\{object\}}}?} \\*
 & (2) \textbf{What does \textnormal{\texttt{\{subject\}}} do with \textnormal{\texttt{\{object\}}}?} \\*
 & (3) \textbf{Which action involving \textnormal{\texttt{\{object\}}} is performed by \textnormal{\texttt{\{subject\}}}?} \\*
 & (4) \textbf{How does \textnormal{\texttt{\{subject\}}} act on or use \textnormal{\texttt{\{object\}}}?} \\*
 & (5) \textbf{What is \textnormal{\texttt{\{subject\}}} observed doing with \textnormal{\texttt{\{object\}}}?} \\*
 & (6) \textbf{Which action connects \textnormal{\texttt{\{subject\}}} with \textnormal{\texttt{\{object\}}} in this interval?} \\*
 & (7) \textbf{What action does \textnormal{\texttt{\{subject\}}} perform?} \\*
 & (8) \textbf{What does \textnormal{\texttt{\{subject\}}} do in the referenced interval?} \\*
 & (9) \textbf{Which action is performed by \textnormal{\texttt{\{subject\}}}?} \\*
 & (10) \textbf{What is \textnormal{\texttt{\{subject\}}} observed doing?} \\*
 & (11) \textbf{How does \textnormal{\texttt{\{subject\}}} act in this part of the video?} \\*
 & (12) \textbf{Which action does the video show \textnormal{\texttt{\{subject\}}} performing?} \\
\cmidrule(lr){1-2}
\multirow[c]{6}{*}{\textbf{TRH}} & (1) \textbf{What is the temporal order between \textnormal{\texttt{\{event\_a\}}} and \textnormal{\texttt{\{event\_b\}}}?} \\*
 & (2) \textbf{Which occurs first: \textnormal{\texttt{\{event\_a\}}} or \textnormal{\texttt{\{event\_b\}}}?} \\*
 & (3) \textbf{How are \textnormal{\texttt{\{event\_a\}}} and \textnormal{\texttt{\{event\_b\}}} ordered in time?} \\*
 & (4) \textbf{What sequence does the video show for \textnormal{\texttt{\{event\_a\}}} and \textnormal{\texttt{\{event\_b\}}}?} \\*
 & (5) \textbf{Does \textnormal{\texttt{\{event\_a\}}} occur before or after \textnormal{\texttt{\{event\_b\}}}?} \\*
 & (6) \textbf{Which temporal relationship holds between \textnormal{\texttt{\{event\_a\}}} and \textnormal{\texttt{\{event\_b\}}}?} \\
\multirow[c]{6}{*}{\textbf{CPH}} & (1) \textbf{What camera or editing change occurs during \textnormal{\texttt{\{camera\_event\}}}?} \\*
 & (2) \textbf{How does the shot actually change during \textnormal{\texttt{\{camera\_event\}}}?} \\*
 & (3) \textbf{Which observed camera or editing operation occurs during \textnormal{\texttt{\{camera\_event\}}}?} \\*
 & (4) \textbf{What camera behavior is visible during \textnormal{\texttt{\{camera\_event\}}}?} \\*
 & (5) \textbf{How is the camera or edit handled during \textnormal{\texttt{\{camera\_event\}}}?} \\*
 & (6) \textbf{Which shot-level operation does the video show during \textnormal{\texttt{\{camera\_event\}}}?} \\
\bottomrule
\end{tabularx}

\vspace{0.30em}

\begin{tabularx}{\textwidth}{@{}>{\centering\arraybackslash}p{0.07\textwidth}>{\raggedright\arraybackslash}X@{}}
\multicolumn{2}{c}{\textbf{Caption task}} \\[2.5pt]
\toprule
\textbf{\mbox{Scope}} & \textbf{\mbox{Exact source templates}} \\
\midrule
\multirow{8}{*}{\textbf{All}} & (1) \textbf{State one directly observable fact from the video in one complete sentence.} \\
 & (2) \textbf{Describe one fact that is directly observable in the video.} \\
 & (3) \textbf{In one complete sentence, report a fact directly supported by the video.} \\
 & (4) \textbf{What is one directly observable fact in the video? Answer in a complete sentence.} \\
 & (5) \textbf{Provide one complete-sentence description of a fact visible or audible in the video.} \\
 & (6) \textbf{Identify one fact directly supported by the video and state it in one sentence.} \\
 & (7) \textbf{Report one concrete observation from the video as a complete sentence.} \\
 & (8) \textbf{Give one complete sentence describing something the video directly establishes.} \\[1pt]
\bottomrule
\end{tabularx}
\endgroup
\end{table*}

\subsection{Communication Protocol}\label{app:communication_protocol}
\looseness=3 To ensure seamless orchestration across the pipeline, \VideoHALO{} implements a structured communication protocol governing both inter-task transitions and intra-task agent collaborations. Within collaborative stages such as FER and GVP, agents exchange intermediate outputs through standardized, schema-driven forms rather than open-ended dialogue. This organization effectively mitigates context dilution and aligns with advanced multi-agent paradigms that coordinate specialized roles via predefined interaction patterns~\cite{hong2024metagpt,wu2024autogen,zhong2026aiharnessengineering}. A fundamental principle of this protocol is the preservation of task state. The propagated information encapsulates the current operational context as well as the verified evidence retained from prior stages. Once rigorously accepted, crucial fields are locked as immutable contextual states. Downstream agents evaluate whether this propagated information provides sufficient support, but they are restricted from overwriting previously validated conclusions. By decoupling localized task reasoning from global state propagation, the protocol guarantees a stable and consistent interpretation of each sample. Any candidate failing to meet the criteria is immediately discarded, though the \textit{Reflection} or \textit{Monitor Agents} may trigger at most one focused visual re-inspection to recover borderline cases. Figures~\ref{fig:cognitive_memory_1} to~\ref{fig:crv_specification} detail the exact instruction schemas and \mbox{communication formats}.

\subsection{Multi-Agent Collaborative Workflow for Video Hallucination Benchmarking}\label{app:multi_agent_workflow}
Algorithm~\ref{alg:videohalo} outlines the complete pipeline, formalizing agent specifications, task orchestration, state transitions, and final acceptance criteria.

\begin{algorithm*}[!t]
\small
\caption{Multi-Agent Collaborative Workflow for Video Hallucination Benchmarking}
\label{alg:videohalo}
\begin{algorithmic}[1]
\setlength{\itemsep}{1.2pt}
\REQUIRE Candidate videos \(\mathcal{V}\), task set \(\mathcal{T}\), memory formulation \(\mathcal{M}=(\mathcal{M}_{\mathrm{cog}},\mathcal{M}_{\mathrm{cat}})\)
\REQUIRE Agents \(\mathcal{A}=\{a_p,a_e,a_r,a_g,a_v,a_m\}\)
\ENSURE Accepted benchmark samples \(\mathcal{D}\)
\STATE \(\mathcal{D}\leftarrow\varnothing\)
\STATE \(\Pi_a\leftarrow\mathcal{M} \oplus \langle\mathcal{R}_a,\mathcal{G}_a,\mathcal{P}_a,\mathcal{O}_a\rangle\quad\forall a\in\mathcal{A}\) \hfill \textcolor{black}{// Initialize multi-agent instructions}
\FOR{each video \(v\in\mathcal{V}\) and task \(\tau\in\mathcal{T}\)}
    \STATE \(\Omega\leftarrow\mathcal{I}(a_p,\Pi_p,\langle v,\tau\rangle)\) \hfill \textcolor{black}{// Retrieve counterfactual feasibility}
    \STATE \(\mu_1\leftarrow\Psi_1(\tau,\Omega,V_1(\Omega))\) \hfill \textcolor{black}{// Stage 1 state checkpoint}
    \STATE \textbf{if} \(\neg\delta^+(\mu_1)\) or \(\operatorname{Constructible}(\Omega)=\varnothing\) \textbf{then continue}
    \STATE \(\mathcal{F}\leftarrow\mathcal{I}(a_e,\Pi_e,\langle v,\mu_1,\operatorname{Constructible}(\Omega)\rangle)\)
    \STATE \textbf{if} \(\mathcal{F}=\varnothing\) \textbf{then continue}
    \FOR{each proposed fact \(f\in\mathcal{F}\)}
        \STATE \(\omega\leftarrow\Gamma(f,\Omega)\) \hfill \textcolor{black}{// Ground text fact to visual matrix}
        \STATE \(q_{\omega}\leftarrow\langle\omega.t,\omega.\alpha,\omega.h\rangle\) \hfill \textcolor{black}{// Time scope, semantic referents, and targeted factual detail}
        \STATE \(r\leftarrow\mathcal{I}(a_r,\Pi_r,\langle v,f,\omega,q_{\omega}\rangle)\)
        \STATE \textbf{if} \(\operatorname{Insufficient}(r)\land\operatorname{Recoverable}(r)\) \textbf{then} \(r\leftarrow\mathcal{I}(a_r,\Pi_r,\langle v,f,\omega,q_{\omega},\mathrm{focused}\rangle)\) \hfill \textcolor{black}{// Targeted visual re-inspection}
        \STATE \(\mu_2\leftarrow\Psi_2(\tau,\omega.c,\omega.e,q_{\omega},f,r,V_2(f,r))\) \hfill \textcolor{black}{// Stage 2 state checkpoint}
        \STATE \textbf{if} \(\neg\delta^+(\mu_2)\) \textbf{then} \(\delta^-(\mu_2)\)\textbf{, continue}
        \STATE \(p\leftarrow\mathcal{I}(a_g,\Pi_g,\langle\mu_2,\tau,q_{\omega}\rangle)\)
        \STATE \(b\leftarrow\mathcal{I}(a_v,\Pi_v,\langle p,f,q_{\omega}\rangle)\) \hfill \textcolor{black}{// Structural back-parsing}
        \STATE \(\mu_3\leftarrow\Psi_3(\tau,\omega.c,\omega.e,q_{\omega},p,b,V_3(p,b,f,q_{\omega}))\) \hfill \textcolor{black}{// Stage 3 state checkpoint}
        \STATE \textbf{if} \(\neg\delta^+(\mu_3)\) \textbf{then} \(\delta^-(\mu_3)\)\textbf{, continue}
        \STATE \(m\leftarrow\mathcal{I}(a_m,\Pi_m,\langle v,\mu_3,f,q_{\omega}\rangle)\)
        \STATE \textbf{if} \(\operatorname{Insufficient}(m)\land\operatorname{Recoverable}(m)\) \textbf{then} \(m\leftarrow\mathcal{I}(a_m,\Pi_m,\langle v,\mu_3,f,q_{\omega},\mathrm{focused}\rangle)\)
        \STATE \(\mu_4\leftarrow\Psi_4(\tau,\omega.c,\omega.e,q_{\omega},f,\mu_3,m,V_4(f,\mu_3,m,q_{\omega}))\) \hfill \textcolor{black}{// Stage 4 state checkpoint}
        \STATE \textbf{if} \(\delta^+(\mu_4)\) \textbf{then} \(\mathcal{D}\leftarrow\mathcal{D}\cup\{\mathcal{S}(v,\tau,\omega,f,\mu_3)\}\) \textbf{else} \(\delta^-(\mu_4)\) \hfill \textcolor{black}{// Commit valid sample}
    \ENDFOR
\ENDFOR
\RETURN \(\mathcal{D}\)
\end{algorithmic}
\end{algorithm*}

\vspace{2pt}\textbf{Algorithm~\ref{alg:videohalo} Notation}\equaldash{---}Let $\mathcal{V}$, $\mathcal{T}$, $\mathcal{A}$, and $\mathcal{D}$ denote the video space, target tasks, multi-agent set, and finalized benchmark dataset, respectively. The system is governed by a unified dual-layer memory formulation $\mathcal{M}=(\mathcal{M}_{\mathrm{cog}}, \mathcal{M}_{\mathrm{cat}})$. For each agent $a \in \mathcal{A}$, the instruction profile is instantiated as $\Pi_a = \mathcal{M} \oplus \langle \mathcal{R}_a, \mathcal{G}_a, \mathcal{P}_a, \mathcal{O}_a \rangle$, integrating the unified memory with localized configurations for agent-specific role, objective, execution process, and output schema. During initialization, $\Omega$ represents the counterfactual feasibility matrix. Given a proposed atomic fact $f \in \mathcal{F}$, the projection operator $\omega = \Gamma(f, \Omega)$ isolates the associated visual context. To maintain state consistency, the invariant tuple $q_{\omega}=\langle\omega.t, \omega.\alpha, \omega.h\rangle$ bounds the time scope, semantic referents, and the targeted factual detail for alteration. Throughout the pipeline, discrete variables $r$, $p$, $b$, and $m$ track the intermediate outputs of reflection, adversarial generation, structural back-parsing, and visual monitoring, respectively. For each stage $k \in \{1, 2, 3, 4\}$, the encapsulation operator $\Psi_k$ consolidates outputs conditioned on a Boolean validity indicator $V_k \in \{0, 1\}$. The resulting checkpoint $\mu_k$ preserves the conceptual category $\omega.c$ and evidence interval $\omega.e$ to prevent subsequent modifications. Operationally, $\mathcal{I}$ executes agent inference, $\delta^+$ and $\delta^-$ act as deterministic gates for stage-level transitions, and the mapping $\mathcal{S}$ projects verified candidates into $\mathcal{D}$.

\begin{figure*}[!t]
\centering
\noindent\makebox[\linewidth][c]{%
\parbox{\linewidth}{%
\hrule
\vspace{0.14em}
{\raggedright\footnotesize\sffamily\bfseries CATEGORICAL MEMORY LAYER\par}
\vspace{0.14em}
\hrule}}\par
\vspace{0.04em}
\includegraphics[width=\linewidth]{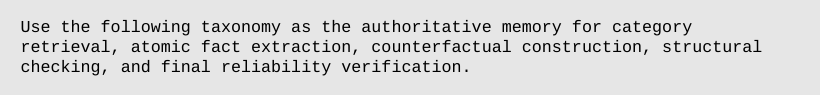}\par
\vspace{-0.3em}
\includegraphics[width=\linewidth]{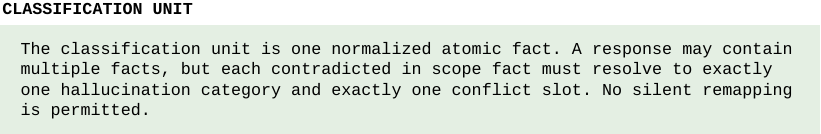}\par
\vspace{-0.3em}
\includegraphics[width=\linewidth]{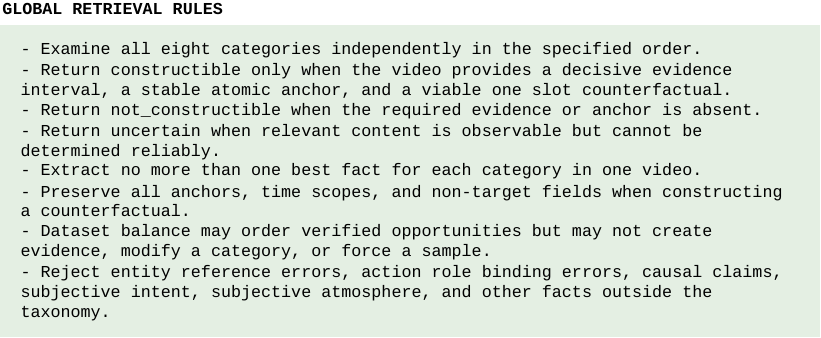}\par
\hrule height 0.4pt\relax
\phantomcaption
\end{figure*}

\begin{figure*}[!t]
\centering
\ContinuedFloat
\noindent\rule{\linewidth}{0.4pt}\par
\includegraphics[width=\linewidth]{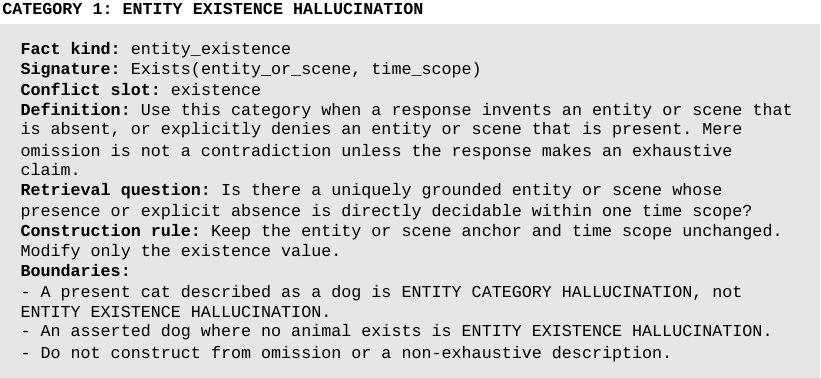}\par
\vspace{-0.3em}
\includegraphics[width=\linewidth]{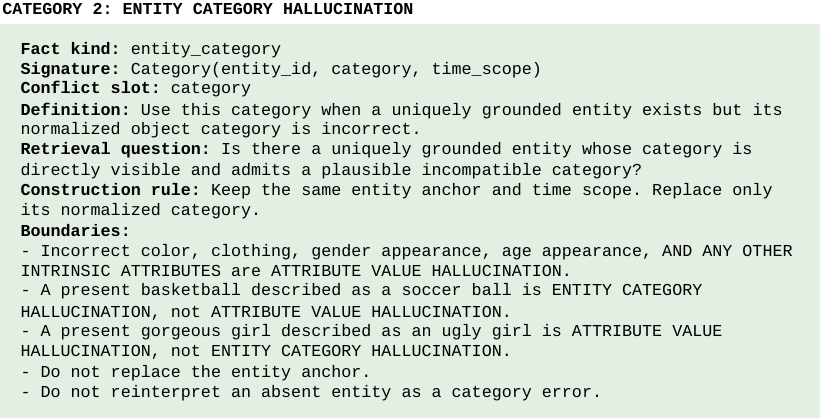}\par
\hrule height 0.4pt\relax
\caption{Categorical Memory Used by \VideoHALO{}, Part I\equaldash{---}This part emphasizes the global category contract, including EEH, ECH and the corresponding exclusion boundaries.}
\label{fig:cognitive_memory}
\end{figure*}

\begin{figure*}[!t]
\centering
\noindent\makebox[\linewidth][c]{%
\parbox{\linewidth}{%
\hrule
\vspace{0.14em}
{\raggedright\footnotesize\sffamily\bfseries CATEGORICAL MEMORY LAYER\par}
\vspace{0.14em}
\hrule}}\par
\vspace{0.04em}
\includegraphics[width=\linewidth]{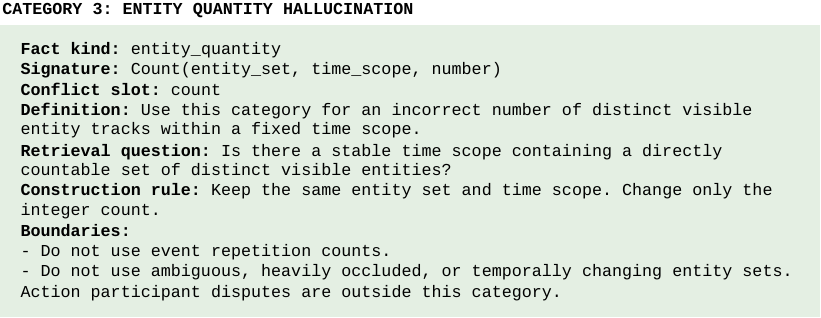}\par
\hrule height 0.4pt\relax
\phantomcaption
\end{figure*}

\begin{figure*}[!t]
\centering
\ContinuedFloat
\noindent\rule{\linewidth}{0.4pt}\par
\includegraphics[width=\linewidth]{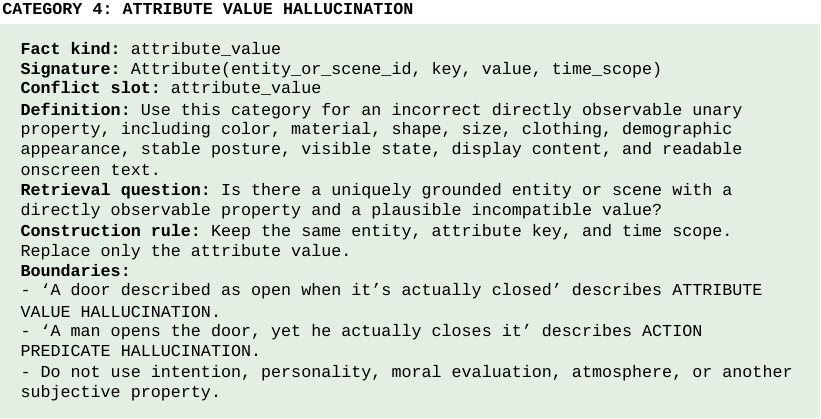}\par
\vspace{-0.3em}
\includegraphics[width=\linewidth]{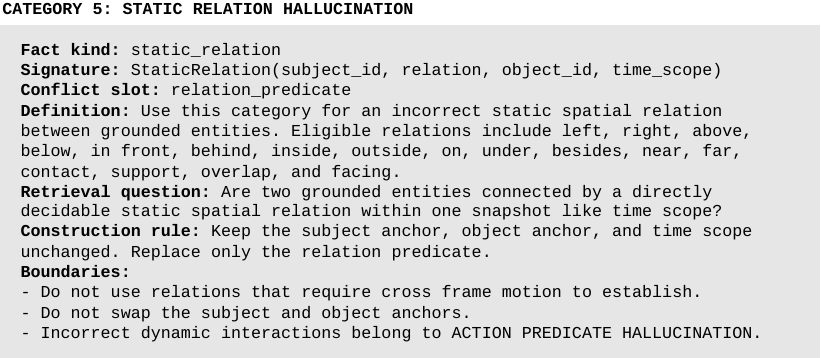}\par
\hrule height 0.4pt\relax
\caption{Categorical Memory Used by \VideoHALO{}, Part II\equaldash{---}This part shows EQH, AVH, and SRH together with the regarding exclusion boundaries.}
\label{fig:category_memory_1}
\end{figure*}

\begin{figure*}[!t]
\centering
\noindent\makebox[\linewidth][c]{%
\parbox{\linewidth}{%
\hrule
\vspace{0.14em}
{\raggedright\footnotesize\sffamily\bfseries CATEGORICAL MEMORY LAYER\par}
\vspace{0.14em}
\hrule}}\par
\vspace{0.04em}
\includegraphics[width=\linewidth]{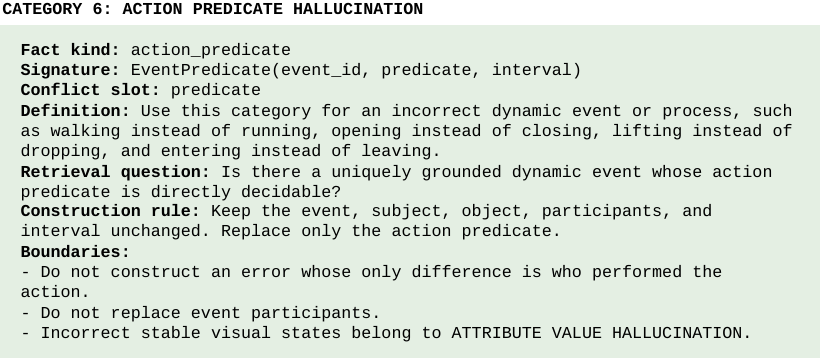}\par
\hrule height 0.4pt\relax
\phantomcaption
\end{figure*}

\begin{figure*}[!t]
\centering
\ContinuedFloat
\noindent\rule{\linewidth}{0.4pt}\par
\includegraphics[width=\linewidth]{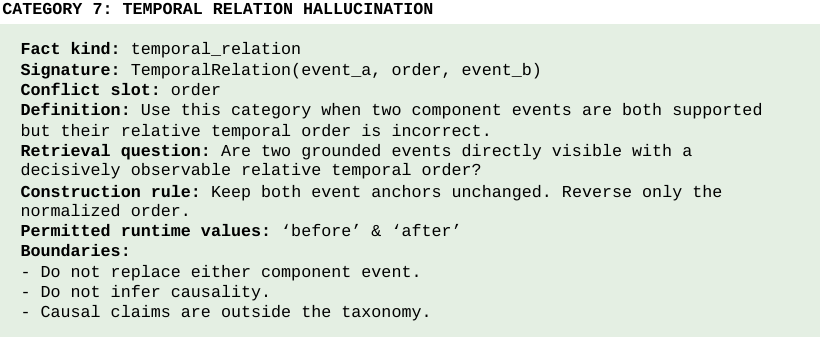}\par
\vspace{-0.3em}
\includegraphics[width=\linewidth]{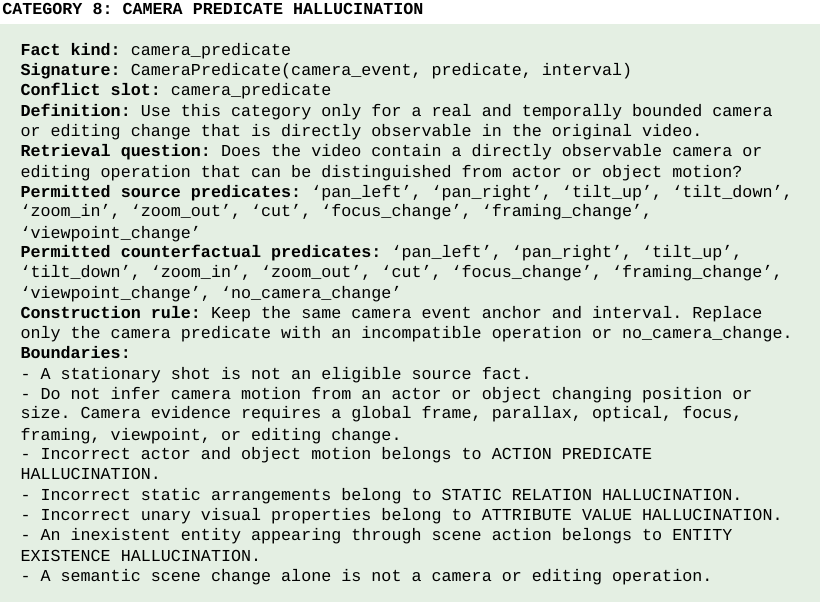}\par
\hrule height 0.4pt\relax
\caption{Categorical Memory Used by \VideoHALO{}, Part III\equaldash{---}This part defines APH, TRH, and CPH with the corresponding exclusion boundaries.}
\label{fig:category_memory_2}
\end{figure*}

\begin{figure*}[!t]
\centering
\setlength{\parskip}{0pt}
\par\hrule\nobreak\vspace{0.5pt}\nobreak
\taskband{taskgreen}{Task:}{Retrieve possible hallucination opportunities from a video.}
\taskplain{Provided Information:}{%
  \taskitem{1}{The complete original video.}
  \taskitem{2}{Task type, either \texttt{Video QA task} or \texttt{Caption task}.}}
\taskband{taskblue}{Goal:}{Identify category-specific visual evidence, a stable atomic anchor, and one viable conflict slot for a controlled counterfactual.}
\nobreak\vspace{1pt}\nobreak
\taskplain{Process:}{%
  \taskitem{1}{The Planner Agent inspects the complete video.}
  \taskitem{2}{Examine all eight categories in the prescribed order.}
  \taskitem{3}{Mark each category as \texttt{constructible}, \texttt{not\_constructible}, or \texttt{uncertain}.}
  \taskitem{4}{Retain evidence intervals and anchor summaries only for constructible categories.}
  \taskitem{5}{Check category coverage, decision state, evidence intervals, atomic anchors, and record structure.}}
\nobreak\vspace{1pt}\nobreak
\taskband{taskyellow}{Output:}{%
  \taskfield{video\_id}{identifier of the inspected video.}
  \taskfield{opportunities}{one typed record for every hallucination category.}
  \taskfield{category\_label}{the examined category.}
  \taskfield{fact\_kind}{the corresponding normalized atomic fact type.}
  \taskfield{conflict\_slot}{the only field that may be changed downstream.}
  \taskfield{constructibility}{\texttt{constructible}, \texttt{not\_constructible}, or \texttt{uncertain}.}
  \taskfield{evidence\_intervals}{decisive temporal evidence for constructible categories.}
  \taskfield{anchor\_summary}{stable entities, events, relations, or camera anchors.}
  \taskfield{decision\_reason}{the evidence-based reason for the decision.}}
\nobreak\vspace{0.5pt}\nobreak\hrule\par
\caption{Hallucination Category Retrieval\equaldash{---}The objective of this sub-task is to scan the video to identify segments suitable for constructing corresponding hallucination types, ultimately returning the category and the decision rationale.}
\label{fig:hcr_specification_task}
\label{fig:hcr_specification}
\end{figure*}

\begin{figure*}[!t]
\centering
\setlength{\parskip}{0pt}
\par\hrule\nobreak\vspace{0.5pt}\nobreak
\taskband{taskgreen}{Task:}{Extract and independently verify grounded atomic facts for the accepted opportunities.}
\taskplain{Provided Information:}{%
  \taskitem{1}{The complete original video.}
  \taskitem{2}{The accepted HCR record, including category, fact kind, conflict slot, evidence intervals, and canonical anchors.}}
\taskband{taskblue}{Goal:}{Produce visually supported atomic facts with correct category assignment, unique grounding, and a viable conflict slot.}
\taskplain{Process:}{%
  \taskitem{1}{The Extraction Agent reinspects each supplied evidence interval.}
  \taskitem{2}{Propose at most one normalized fact for each constructible opportunity.}
  \taskitem{3}{Preserve category, fact kind, conflict slot, time scope, and canonical anchors.}
  \taskitem{4}{The Reflection Agent independently rereads the original video.}
  \taskitem{5}{Verify support, unique grounding, category boundaries, and mutation viability.}
  \taskitem{6}{A recoverable insufficient verdict permits one focused rereading. Unresolved records are excluded.}}
\nobreak\vspace{1pt}\nobreak
\taskband{taskyellow}{Output:}{%
  \taskfield{verified\_factbank}{the accepted typed fact records.}
  \taskfield{source\_fact\_id}{identifier of the proposed fact.}
  \taskfield{fact\_kind}{one of the eight normalized fact kinds.}
  \taskfield{natural\_language\_fact}{the complete factual statement.}
  \taskfield{time\_scope}{the supporting temporal interval.}
  \taskfield{normalized\_fact}{the structured atomic fact.}
  \taskfield{reflection\_report}{verdict, unique grounding, category validity, mutation viability, evidence interval, evidence summary, and recoverability.}}
\nobreak\vspace{0.5pt}\nobreak\hrule\par
\caption{Fact Extraction and Reflection\equaldash{---}The objective of this sub-task is to inherit the output fields from the preceding sub-task under a uniform communication protocol, map the segments back to their corresponding video intervals for fact extraction, and verify the reliability of these extracted facts through independent reflection.}
\label{fig:fer_specification_task}
\label{fig:fer_specification}
\end{figure*}

\begin{figure*}[!t]
\centering
\setlength{\parskip}{0pt}
\par\hrule\nobreak\vspace{0.5pt}\nobreak
\taskband{taskgreen}{Task:}{Generate and structurally verify adversarial items from a textual record without video access.}
\taskplain{Provided Information:}{%
  \taskitem{1}{The sealed FER record and its verified natural-language fact.}
  \taskitem{2}{Task type, either \texttt{Video QA task} or \texttt{Caption task}.}
  \taskitem{3}{One declared mutation operator for the selected category.}
  \taskitem{4}{A fixed question template and the canonical anchors that must remain unchanged.}}
\taskband{taskblue}{Goal:}{Construct factual and counterfactual responses that differ only at the declared conflict slot.}
\nobreak\vspace{1pt}\nobreak
\taskplain{Process:}{%
  \taskitem{1}{Use the fixed question selected from the verified fact and task template.}
  \taskitem{2}{The Generation Agent selects a plausible incorrect value for the target slot.}
  \taskitem{3}{Keep the factual response unchanged and alter exactly one slot in the counterfactual response.}
  \taskitem{4}{The text-only Verification Agent back parses both responses into normalized facts.}
  \taskitem{5}{A deterministic check compares both reconstructions with the sealed fact and declared conflict slot.}
  \taskitem{6}{Advance only if the question and protected fields match and exactly one target slot differs.}}
\nobreak\vspace{1pt}\nobreak
\taskband{taskyellow}{Output:}{%
  \taskfield{replacement\_value}{the selected false value.}
  \taskfield{question}{the fixed question supplied by the selected template.}
  \taskfield{answer}{the factual response.}
  \taskfield{counterfactual\_answer}{the adversarial response.}
  \taskfield{supported\_fact}{the reconstructed factual structure.}
  \taskfield{counterfactual\_fact}{the reconstructed adversarial structure.}
  \taskfield{graph\_diff}{one changed atomic fact, one changed slot, and only the declared conflict path.}}
\nobreak\vspace{0.5pt}\nobreak\hrule\par
\caption{Generation and Verification of Adversarial Pairs\equaldash{---}The objective of this sub-task is to construct adversarial examples for the corresponding video understanding tasks. It first inherits the output fields from the preceding sub-task to inject structured facts into predefined question templates. Simultaneously, it generates counterfactual statements based on these facts and utilizes reverse parsing to ensure the structural correctness of the sample.}
\label{fig:gvp_specification}
\end{figure*}

\begin{figure*}[!t]
\centering
\setlength{\parskip}{0pt}
\par\hrule\nobreak\vspace{0.5pt}\nobreak
\taskband{taskgreen}{Task:}{Validate the reliability of the complete sample.}
\taskplain{Provided Information:}{%
  \taskitem{1}{The complete original video.}
  \taskitem{2}{The sealed GVP sample with its question, factual response, and counterfactual response.}
  \taskitem{3}{The authoritative verified fact, time scope, canonical anchors, and conflict slot.}}
\taskband{taskblue}{Goal:}{Accept samples exclusively when factual responses are supported and counterfactual responses are contradicted alongside correct categories and zero additional errors.}
\taskplain{Process:}{%
  \taskitem{1}{The Monitor Agent independently reinspects the complete original video.}
  \taskitem{2}{Verify the factual response against the authoritative source fact.}
  \taskitem{3}{Verify that the counterfactual response is contradicted by the video.}
  \taskitem{4}{Check category boundaries, grounding, the one-slot constraint, and additional errors.}
  \taskitem{5}{A recoverable insufficient verdict permits one focused rereading. Unresolved samples are rejected.}
  \taskitem{6}{A deterministic terminal check admits the sealed samples only after the report passes.}}
\nobreak\vspace{1pt}\nobreak
\par\hrule\nobreak\vspace{1.5pt}\nobreak
\taskband{taskyellow}{Output:}{%
  \taskfield{reliability\_report}{acceptance state and factual and counterfactual verdicts.}
  \taskfield{semantic\_checks}{source-fact match, category boundary, planned-category target, one-slot consistency, and additional error count.}
  \taskfield{evidence}{the decisive interval, evidence summary, and recoverable reason.}
  \taskfield{accepted\_record}{benchmark-facing fields created only after terminal acceptance.}}
\par\nobreak\vspace{0.5pt}\nobreak\hrule\par\nobreak
\caption{Comprehensive Reliability Validation\equaldash{---}The objective of this sub-task is to review the constructed adversarial examples, ensuring that the generated hallucination types align with expectations. Specifically, the factual and counterfactual answers must contradict each other exclusively regarding the targeted hallucination while maintaining consistent overall structures.}
\label{fig:crv_specification}
\end{figure*}

\section{External Human Audit}\label{app:human_verification}
\paragraph{Sampling and Reviewers}\equaldash{\textemdash}We randomly sampled 100 instances from each of the eight original categories, yielding an audit set of 800 samples from the 2,000-sample benchmark. Sampling used a stable sort and seed 20260818. The subset contains 649 unique videos, with 413 Video Question Answering (Video QA) and 387 Video Captioning samples. Two workers independently reviewed the subset after a 30-sample calibration phase. One author independently reviewed all 800 samples to establish the reference category labels.

\paragraph{Annotation Protocol}\equaldash{\textemdash}Workers were provided with the video, question, candidate answers, and category definitions. Original category labels and the other worker's judgments were hidden, and presentation order was randomized independently for each worker. Workers verified that the factual answer was supported by the video while the counterfactual answer was contradicted by it, and that the two answers differed only in the targeted content. Each sample was assigned a final category label.

\paragraph{Metrics}\equaldash{\textemdash}Sample accuracy is the proportion of original benchmark categories matching the author's reference labels. Worker agreement is the proportion of identical category labels assigned by the two workers. Both agreement and Cohen's $\kappa$ use the workers' independent annotations. We retain $\texttt{InvalidSample}$ as a distinct label prior to modification. Per-category accuracy and agreement are evaluated within each 100-sample stratum. Given the equal strata sizes, the overall values across all 800 samples naturally equal their macro-averages. Additionally, per-category $\kappa$ uses one-vs-rest coding over the full 800 samples, while overall $\kappa$ relies on the complete multiclass labels.

\paragraph{Uncertainty}\equaldash{\textemdash}We estimated 95\% percentile confidence intervals using 10,000 stratified bootstrap replicates with seed 20260818. Each replicate sampled 100 items with replacement within each original category, retaining the reference and both workers' labels for each sampled item. All metrics were recomputed in every replicate, and intervals were defined by the 2.5th and 97.5th percentiles (Table~\ref{tab:human_audit_ci}). We used NumPy's PCG64 generator, ordered categories as EEH, ECH, EQH, AVH, SRH, APH, TRH, and CPH, and sorted sample IDs lexicographically within each category.
\begingroup
\footnotesize
\renewcommand{\arraystretch}{1.12}
\setlength{\LTleft}{0pt plus 1fill}
\setlength{\LTright}{0pt plus 1fill}
\setlength{\LTcapwidth}{\textwidth}
\setlength{\LTpre}{2pt}
\setlength{\LTpost}{2pt}
\begin{table*}[!t]
\centering
\footnotesize
\setlength{\tabcolsep}{2.6pt}
\renewcommand{\arraystretch}{1.06}

\begin{tabularx}{\textwidth}{@{}>{\raggedright\arraybackslash}p{0.09\textwidth}*{3}{>{\centering\arraybackslash}X}@{}}
\toprule
\textbf{\mbox{Scope}} & \textbf{\mbox{Sample accuracy}} & \textbf{\mbox{Worker agreement}} & \textbf{\mbox{Cohen's $\kappa$}} \\
\midrule
EEH & 0.9900 [0.9700, 1.0000] & 0.9700 [0.9300, 1.0000] & 0.9826 [0.9588, 1.0000] \\
ECH & 0.9300 [0.8800, 0.9800] & 0.9600 [0.9200, 0.9900] & 0.9752 [0.9484, 0.9941] \\
EQH & 1.0000 [1.0000, 1.0000] & 1.0000 [1.0000, 1.0000] & 1.0000 [1.0000, 1.0000] \\
AVH & 1.0000 [1.0000, 1.0000] & 0.9800 [0.9500, 1.0000] & 0.9627 [0.9337, 0.9890] \\
SRH & 1.0000 [1.0000, 1.0000] & 0.9900 [0.9700, 1.0000] & 0.9886 [0.9713, 1.0000] \\
APH & 0.9800 [0.9500, 1.0000] & 0.9600 [0.9200, 0.9900] & 0.9763 [0.9513, 0.9943] \\
TRH & 1.0000 [1.0000, 1.0000] & 0.9300 [0.8800, 0.9700] & 0.9580 [0.9242, 0.9826] \\
CPH & 1.0000 [1.0000, 1.0000] & 0.9400 [0.8900, 0.9800] & 0.9645 [0.9334, 0.9885] \\*
\midrule
\textbf{All} & \textbf{0.9875 [0.9800, 0.9950]} & \textbf{0.9663 [0.9525, 0.9788]} & \textbf{0.9616 [0.9461, 0.9758]} \\
\bottomrule
\end{tabularx}
\caption{Category Annotation Audit with 95\% Stratified Bootstrap Confidence Intervals --- All values range from 0 to 1. Accuracy compares original benchmark categories against the author reference labels while agreement and $\kappa$ assess consensus between the two workers. Accuracy and agreement for each category rely on the original sampling strata. The $\kappa$ metric per category applies one versus rest coding across all 800 instances and the overall column reports the multiclass $\kappa$.}
\label{tab:human_audit_ci}
\end{table*}
\endgroup

\section{Main Experiment Settings}\label{app:main_experiment_settings}

\paragraph{Evaluation Subset and Split}\equaldash{\textemdash}We selected 100 instances from each of the eight categories in the \VidHalLoc{} benchmark of 2,000 examples, yielding an evaluation set of 800 items spanning 436 unique videos. The calibration set contains 100 instances from 100 distinct videos, ensuring zero overlap with the evaluation data.

\subsection{Hardware and Dependency Settings}\label{app:hardware_dependencies}

\paragraph{Commercial LVLMs}\equaldash{\textemdash}Following local video pre\-processing, Gemini-3-Flash~\cite{google2025gemini3flash} and GPT-5~\cite{openai2025gpt5} were accessed via their respective provider APIs. Specifically, Gemini-3-Flash utilized the \texttt{gemini-\allowbreak3-\allowbreak flash-\allowbreak preview} endpoint to handle provider-native video inputs, whereas GPT-5 was supplied with frames uniformly sampled at 1 fps. 

\paragraph{Open-source LVLMs}\equaldash{\textemdash}A diverse suite of open-source models\equaldash{—}LLaVA-NeXT-Video~\cite{li2024llavanextinterleave}, Qwen3-VL-Instruct~\cite{bai2025qwen3vl}, Gemma-4-it~\cite{gemmateam2026gemma4}, InternVL3.5~\cite{wang2025internvl35}, and Qwen3-Omni-Instruct~\cite{xu2025qwen3omni}\equaldash{—}were deployed locally on a cluster of four A100-PCIE-40GB GPUs. The standardized software environment comprised Python 3.12.3, PyTorch 2.8.0+cu128, TorchVision 0.23.0+cu128, and Transformers 5.14.1, with InternVL3.5 acting as the sole exception requiring Transformers 4.52.1. In terms of precision, LLaVA-NeXT-Video operated in FP16, while the remainder utilized BF16. All models processed 32-frame inputs; notably, InternVL3.5 extracted segment midpoints, and Qwen3-Omni-Instruct was evaluated with its audio and talker modules explicitly disabled. Furthermore, Qwen3.6-27B~\cite{qwenteam2026qwen36} was distributed across the same four-GPU hardware setup, standardizing on 32 frames and BF16 precision, but operated within a specialized environment featuring NVIDIA PyTorch 2.7.0a0+nv25.03 and Transformers 5.14.1. 

\paragraph{Video Agents}\equaldash{\textemdash}We evaluated several agentic frameworks. VideoAgent~\cite{fan2024videoagentmemory} integrated a GPT-4o~\cite{openai2024gpt4o} controller with a local 4-bit Video-LLaVA-7B~\cite{lin2024videollava} model, LaViLa~\cite{zhao2023lavila}, and an object memory module to facilitate adaptive video access. Both VideoHV-Agent~\cite{wang2026think} and Deep Video Discovery (DVD)~\cite{zhang2025deepvideodiscovery} relied on open-source agent implementations backed by provider-side GPT-4o inference. Specifically, VideoHV-Agent (Python 3.10.20) maintained a 1 fps frame library capped at 180 frames, whereas DVD (Python 3.11.15) applied 2 fps sampling across 10~s clips, leveraging GPT-4o for orchestration alongside \texttt{text-embedding-3-large}~\cite{openai2024embedding3}. 

\paragraph{Detection Methods}\equaldash{\textemdash}For hallucination detection, PAC-S~\cite{sarto2023positive} and EMScore~\cite{shi2021emscore} evaluated all decoded frames locally on the aforementioned A100-PCIE-40GB pool. PAC-S operated under Python 3.9.16 and PyTorch 1.12.1+cu113 utilizing its dedicated CLIP ViT-B/32 checkpoint, whereas EMScore ran on Python 3.8.20 and PyTorch 1.7.1+cu110 using the standard OpenAI CLIP ViT-B/32~\cite{radford2021clip}. Owl-Con~\cite{bansal2024videocon} was similarly deployed on this GPU pool, processing 32 frames via mPLUG-Owl-LLaMA-7B-Video~\cite{ye2023mplugowl} (Owl-Con checkpoint) within an environment comprising PyTorch 1.13.1+cu117, TorchVision 0.14.1+cu117, Transformers 4.28.1, and PEFT 0.4.0. Alternatively, FIFA~\cite{jing2025fifa} adopted a hybrid methodology, combining local Qwen2.5-VL-72B~\cite{bai2025qwen25vl} inference on two A100-SXM4-80GB GPUs with GPT-4o~\cite{openai2024gpt4o} for DSG. Its specialized local environment leveraged PyTorch 2.7.0a0+nv25.3, Transformers 4.57.1, FlashAttention 2.7.3~\cite{dao2022flashattention}, Accelerate 1.6.0, \texttt{qwen-vl-utils} 0.0.14, and Decord 0.6.0~\cite{dmlc2021decord}, sampling at 1 fps for Video QA verification. 

\paragraph{System Configuration}\equaldash{\textemdash}The centralized audit host was equipped with 14 CPU cores, 240~GB RAM, and NVIDIA driver 570.124.06, exposing two 81,920~MiB devices. Standard media I/O operations were consistently handled by FFmpeg~\cite{tomar2006ffmpeg}, OpenCV~\cite{opencv_library}, Decord~\cite{dmlc2021decord}, and PyAV~\cite{pyav2026}, while all external provider API calls were configured as resumable requests to ensure robust execution.

\subsection{Threshold Settings}\label{app:threshold_settings}

For score-based methods, binary predictions ($\hat{y} \in \{0, 1\}$) are derived by mapping candidate scores against a designated threshold. As formalized in Algorithm~\ref{alg:threshold_selection}, we determine a method-specific global threshold optimized exclusively on a disjoint validation split, which is strictly frozen prior to the main evaluation.

\begin{algorithm*}[!t]
\small
\caption{Global Threshold Selection}
\label{alg:threshold_selection}
\begin{algorithmic}[1]
\setlength{\itemsep}{0.8pt}
\STATE $\Theta\leftarrow\mathcal{B}(S_F\cup S_C)$ and $m\leftarrow\tfrac{1}{2}\bigl(\operatorname{median}(S_F)+\operatorname{median}(S_C)\bigr)$
\FOR{each $t\in\Theta$}
    \STATE Yield indicator $\hat{y}=1 \iff s\geq t$, and compute $A_F(t)$, $A_C(t)$, $A_O(t)$
    \STATE $K(t)\leftarrow\left\langle A_O(t),\min\{A_F(t),A_C(t)\},\tfrac{A_F(t)+A_C(t)}{2},-|t-m|,-t\right\rangle$
\ENDFOR
\RETURN $t^\star\leftarrow\arg\max_{t\in\Theta}^{\succ_L}K(t)$ \hfill \textcolor{black}{// Frozen for downstream evaluation}
\end{algorithmic}
\end{algorithm*}

\begin{table*}[!t]
\centering
\footnotesize
\renewcommand{\arraystretch}{1.00}
\begin{tabular*}{\linewidth}{@{\extracolsep{\fill}}lrrrr@{}}
\toprule
\textbf{\mbox{Method}} & \textbf{\mbox{Threshold}} & \textbf{\mbox{Factual}} & \textbf{\mbox{Counterfactual}} & \textbf{\mbox{Overall}} \\
\midrule
EMScore \mbox{\cite{shi2021emscore}} & 0.2649626136 & 66.0\% & 43.0\% & 10.0\% \\
PAC-S \mbox{\cite{sarto2023positive}} & 0.7559156418 & 54.0\% & 56.0\% & 13.0\% \\
Owl-Con \mbox{\cite{bansal2024videocon}} & 0.5312500000 & 66.0\% & 66.0\% & 38.0\% \\
FIFA \mbox{\cite{jing2025fifa}} & 0.8000000000 & 40.0\% & 94.0\% & 36.0\% \\
\bottomrule
\end{tabular*}
\caption{Global Thresholds and Validation Performance\equaldash{---}We use a disjoint validation split of 100 samples to calibrate the decision thresholds for all accuracy evaluations. For the main evaluation results, see Table~\ref{tab:main_results}.}
\label{tab:detector_thresholds}
\end{table*}

\begin{table*}[!t]
\centering
\small
\setlength{\tabcolsep}{3pt}
\renewcommand{\arraystretch}{1.12}
\begin{tabular*}{\textwidth}{@{\extracolsep{\fill}}lcccc@{}}
\toprule
\textbf{\mbox{Method}} & \textbf{\mbox{Overall}} & \textbf{\mbox{AUROC $\times 100$}} & \textbf{\mbox{Ranking}} & \textbf{\mbox{Ties}} \\
\midrule
EMScore & 6.50 [4.89, 8.24] & 52.94 [52.27, 53.67] & 61.75 [58.33, 65.16] & 0.00 \\
PAC-S & 7.25 [5.47, 9.15] & 52.98 [52.30, 53.71] & 60.38 [57.18, 63.52] & 0.00 \\
Owl-Con & 33.13 [29.85, 36.38] & 67.76 [65.60, 69.94] & 70.31 [67.11, 73.43] & 8.63 \\
FIFA & 34.63 [31.26, 38.04] & 66.03 [64.24, 67.95] & 74.94 [72.58, 77.36] & 27.88 \\
\bottomrule
\end{tabular*}
\caption{Raw Score Summaries For the Four Detectors --- AUROC is scaled by 100, and Overall, Ranking, and Ties are percentages. Ties denotes equal factual and counterfactual scores within an instance.}
\label{tab:score_diagnostics}
\end{table*}

\begin{table*}[!t]
\centering
\footnotesize
\setlength{\tabcolsep}{3pt}
\renewcommand{\arraystretch}{1.12}
\begin{tabular*}{\textwidth}{@{\extracolsep{\fill}}llcccc@{}}
\toprule
\textbf{\mbox{Method}} & \textbf{\mbox{Group}} & \textbf{\mbox{AUROC $\times 100$}} & \textbf{\mbox{Ranking}} & \textbf{\mbox{Overall}} & \textbf{\mbox{Ties}} \\
\midrule
EMScore & OH & 54.38 [53.49, 55.37] & 67.00 [62.86, 71.15] & 8.00 [5.71, 10.46] & 0.00 \\
EMScore & DH & 50.74 [49.65, 51.83] & 53.00 [47.25, 58.61] & 4.00 [1.99, 6.35] & 0.00 \\
\addlinespace[2pt]
PAC-S & OH & 54.43 [53.50, 55.47] & 65.60 [61.41, 69.76] & 10.00 [7.39, 12.85] & 0.00 \\
PAC-S & DH & 50.54 [49.58, 51.49] & 51.67 [46.06, 57.24] & 2.67 [1.00, 4.61] & 0.00 \\
\addlinespace[2pt]
Owl-Con & OH & 72.03 [69.07, 74.97] & 71.20 [67.28, 75.00] & 42.40 [38.15, 46.69] & 6.00 \\
Owl-Con & DH & 61.82 [59.03, 64.62] & 68.83 [63.94, 73.59] & 17.67 [13.56, 21.84] & 13.00 \\
\addlinespace[2pt]
FIFA & OH & 71.79 [69.33, 74.31] & 81.70 [78.69, 84.63] & 42.40 [38.05, 46.84] & 17.40 \\
FIFA & DH & 58.27 [55.79, 60.85] & 63.67 [59.79, 67.58] & 21.67 [17.22, 26.28] & 45.33 \\
\bottomrule
\end{tabular*}
\caption{Detector scores for OH (500 instances) and DH (300 instances).}
\label{tab:category_score_ohdh}
\end{table*}

\begin{table*}[!t]
\centering
\footnotesize
\setlength{\tabcolsep}{2.6pt}
\renewcommand{\arraystretch}{1.06}
\setlength{\tabcolsep}{3pt}

\begin{tabularx}{\textwidth}{@{}l c *{3}{>{\centering\arraybackslash}X} r@{}}
\toprule
\textbf{\mbox{Method}} & \textbf{\mbox{Category}} & \textbf{\mbox{AUROC $\times 100$}} & \textbf{\mbox{Ranking}} & \textbf{\mbox{Overall}} & \textbf{\mbox{Ties}} \\
\midrule
EMScore & EEH & 53.32 [51.61, 55.49] & 66.00 [56.44, 75.26] & 4.00 [0.92, 8.24] & 0.00 \\
EMScore & ECH & 59.31 [56.20, 63.11] & 75.00 [66.32, 83.17] & 15.00 [8.33, 22.47] & 0.00 \\
EMScore & EQH & 52.17 [51.00, 53.73] & 66.00 [56.38, 75.22] & 6.00 [1.90, 11.11] & 0.00 \\
EMScore & AVH & 57.08 [55.09, 59.77] & 77.00 [68.54, 85.00] & 8.00 [3.09, 13.69] & 0.00 \\
EMScore & SRH & 50.50 [48.94, 52.15] & 51.00 [41.24, 60.79] & 7.00 [2.41, 12.50] & 0.00 \\
EMScore & APH & 53.43 [50.68, 56.34] & 64.00 [54.17, 73.56] & 8.00 [3.09, 13.68] & 0.00 \\
EMScore & TRH & 48.93 [47.72, 49.88] & 41.00 [31.31, 50.91] & 1.00 [0.00, 3.37] & 0.00 \\
EMScore & CPH & 49.95 [47.67, 52.19] & 54.00 [44.12, 63.72] & 3.00 [0.00, 6.74] & 0.00 \\
\midrule
PAC-S & EEH & 47.56 [45.72, 49.20] & 40.00 [30.48, 49.58] & 1.00 [0.00, 3.41] & 0.00 \\
PAC-S & ECH & 58.76 [55.95, 62.29] & 71.00 [62.04, 79.49] & 16.00 [9.09, 23.58] & 0.00 \\
PAC-S & EQH & 57.64 [55.85, 60.16] & 83.00 [75.31, 90.10] & 13.00 [6.74, 20.00] & 0.00 \\
PAC-S & AVH & 56.25 [54.05, 59.02] & 74.00 [65.12, 82.24] & 15.00 [8.25, 22.34] & 0.00 \\
PAC-S & SRH & 52.24 [50.76, 54.08] & 60.00 [50.00, 69.66] & 5.00 [1.04, 9.78] & 0.00 \\
PAC-S & APH & 51.37 [48.72, 54.04] & 56.00 [46.15, 65.48] & 4.00 [0.92, 8.26] & 0.00 \\
PAC-S & TRH & 49.33 [48.36, 50.14] & 45.00 [34.96, 54.84] & 0.00 [0.00, 0.00] & 0.00 \\
PAC-S & CPH & 50.79 [49.11, 52.58] & 54.00 [44.07, 63.74] & 4.00 [0.90, 8.33] & 0.00 \\
\midrule
Owl-Con & EEH & 88.35 [83.91, 92.62] & 94.00 [89.22, 98.00] & 67.00 [57.61, 76.09] & 2.00 \\
Owl-Con & ECH & 73.67 [67.39, 79.61] & 68.50 [59.31, 77.33] & 40.00 [30.61, 49.55] & 3.00 \\
Owl-Con & EQH & 67.34 [58.20, 75.76] & 63.00 [53.67, 72.16] & 48.00 [38.04, 57.84] & 4.00 \\
Owl-Con & AVH & 70.94 [64.67, 76.91] & 73.00 [64.14, 81.25] & 40.00 [30.53, 49.51] & 6.00 \\
Owl-Con & SRH & 57.90 [53.14, 62.86] & 57.50 [48.57, 66.49] & 17.00 [10.00, 25.00] & 15.00 \\
Owl-Con & APH & 65.25 [59.17, 71.17] & 68.00 [59.04, 76.60] & 21.00 [13.13, 29.03] & 10.00 \\
Owl-Con & TRH & 58.64 [55.42, 62.36] & 73.50 [65.38, 81.18] & 12.00 [6.00, 18.68] & 13.00 \\
Owl-Con & CPH & 62.46 [57.22, 67.82] & 65.00 [56.19, 73.33] & 20.00 [12.50, 28.28] & 16.00 \\
\midrule
FIFA & EEH & 67.92 [61.66, 74.19] & 75.00 [66.98, 82.55] & 39.00 [29.36, 48.94] & 14.00 \\
FIFA & ECH & 73.16 [68.33, 78.06] & 83.00 [76.29, 89.29] & 36.00 [26.53, 45.83] & 14.00 \\
FIFA & EQH & 77.87 [72.92, 83.26] & 88.50 [83.50, 93.06] & 56.00 [46.32, 66.00] & 17.00 \\
FIFA & AVH & 73.78 [68.86, 79.14] & 87.50 [82.00, 92.38] & 44.00 [34.07, 53.54] & 17.00 \\
FIFA & SRH & 68.16 [62.79, 74.09] & 74.50 [67.50, 81.38] & 37.00 [27.78, 46.79] & 25.00 \\
FIFA & APH & 62.09 [57.43, 67.19] & 70.50 [63.55, 77.12] & 31.00 [22.22, 40.40] & 35.00 \\
FIFA & TRH & 55.01 [51.70, 58.85] & 61.50 [54.89, 67.86] & 21.00 [13.25, 29.17] & 51.00 \\
FIFA & CPH & 59.52 [54.19, 65.10] & 59.00 [52.20, 65.71] & 13.00 [6.80, 19.82] & 50.00 \\
\bottomrule
\end{tabularx}
\caption{Detector scores by hallucination category (100 instances per category).}\label{tab:category_score_leaf}
\end{table*}

\vspace{2pt}\textbf{Algorithm~\ref{alg:threshold_selection} Notation}\equaldash{---}Let $S_F$ and $S_C$ denote the multisets of factual and counterfactual scores from the validation split while $s$ represents a given candidate score. The candidate threshold space $\Theta$ is generated by the discrete boundary operator $\mathcal{B}(S_F \cup S_C)$. For any threshold $t \in \Theta$, the binary decision rule yields an indicator $\hat{y} = 1$ if $s \geq t$ and $\hat{y} = 0$ otherwise. The functions $A_F(t)$ and $A_C(t)$ denote the marginal accuracies for factual acceptance ($\hat{y}=1$) and counterfactual rejection ($\hat{y}=0$) while $A_O(t)$ defines the Overall pairwise accuracy for simultaneous correct classifications. The reference midpoint $m$ is defined as the arithmetic mean of the medians of $S_F$ and $S_C$. Algorithm~\ref{alg:threshold_selection} maximizes $K(t)$ lexicographically by using later components only when all preceding components tie. The selected threshold $t^\star$ remains fixed across all evaluation splits and tasks as well as hallucination categories.

Using the stored independent candidate scores for the 800 evaluation instances, Tables~\ref{tab:score_diagnostics} through \ref{tab:category_score_leaf} summarize the four detectors across the full set and OH/DH subsets alongside individual categories for pooled Video QA and Video Captioning. AUROC compares factual and counterfactual score distributions within each group while Ranking denotes the fraction of instances with $s_F>s_C$ by assigning half credit to ties. The Overall metric uses the fixed thresholds in Table~\ref{tab:detector_thresholds}. Brackets provide pointwise 95\% percentile intervals from 10,000 bootstrap resamples of the 436 video IDs (seed 20260906) to retain all instances and methods within each sampled video.

The lower DH AUROCs and variation per category align consistently with the accuracy patterns in Section~\ref{eval_evaluators}.

\section{Statistical Analysis}\label{app:statistical_analysis}
\subsection{Macro-Level Evaluation of Hallucination Categories}\label{app:macro_level_evaluation}

\begin{table*}[!t]
\centering
\footnotesize
\renewcommand{\arraystretch}{1.12}
\renewcommand{\tabularxcolumn}[1]{m{#1}}
\begin{tabularx}{\textwidth}{@{}ccccc>{\centering\arraybackslash}X>{\centering\arraybackslash}X@{}}
\toprule
\textbf{\mbox{Metric}} & \textbf{\mbox{Ontology}} & \textbf{\mbox{Dynamic}} & \textbf{\mbox{$\Delta$}} & \textbf{\mbox{Cluster SE}} & \textbf{\mbox{Cluster 95\% CI}} & \textbf{\mbox{Category 95\% CI}} \\
\midrule
Factual & 76.55 & 70.53 & 6.01 & 1.11 & [3.87, 8.22] & [4.01, 8.07] \\
Counterfactual & 75.32 & 61.47 & 13.85 & 1.23 & [11.41, 16.22] & [11.53, 16.17] \\
Overall & 54.51 & 34.98 & 19.53 & 1.39 & [16.79, 22.28] & [17.03, 22.06] \\
\bottomrule
\end{tabularx}
\caption{Macro-Average Comparison Between Ontology and Dynamic Hallucinations\equaldash{---}Ontology and Dynamic denote accuracy percentages, while $\Delta$ signifies their difference in percentage points. The strict criterion requires both responses within a matched sample to be classified correctly. Cluster SE and Cluster 95\% CI derive from 10,000 bootstrap resamples of 436 video-ID clusters. The Category 95\% CI functions as an aggregate sensitivity interval for $\Delta$, calculated by resampling 100 instances with replacement within each of the distinct categories.}
\label{tab:aggregate_bootstrap}
\end{table*}

\begin{table*}[tp]
\centering
\small
\setlength{\tabcolsep}{4pt}
\renewcommand{\arraystretch}{1.12}
\begin{tabular*}{\textwidth}{@{\extracolsep{\fill}}lrrrrrl@{}}
\toprule
\textbf{\mbox{Subset}} & \textbf{\mbox{Videos}} & \textbf{\mbox{Instances}} & \textbf{\mbox{OH}} & \textbf{\mbox{DH}} & \textbf{\mbox{$\Delta$}} & \textbf{\mbox{95\% CI}} \\
\midrule
Full evaluation & 436 & 800 & 54.51 & 34.98 & 19.53 & [16.79,22.28] \\
Within-video & 198 & 396 & 53.74 & 34.61 & 19.12 & [15.59,22.69] \\
\bottomrule
\end{tabular*}
\caption{Supplementary Comparison Within the Same Video in Table~\ref{tab:aggregate_bootstrap} --- Each of the 198 videos provides one OH and one DH instance of the same task, with averages weighting all videos and the fifteen methods equally. This subset employs the aforementioned bootstrap over video clusters with 10,000 draws (seed 20260906) to retain pairs and methods jointly. The OH and DH metrics report Overall accuracy (\%) while $\Delta$ and its 95\% CI denote percentage points calculated before rounding.}
\label{tab:within_video_comparison}
\end{table*}

The primary analysis compares Ontology and Dynamic hallucinations. The Ontology category comprises EEH, ECH, EQH, AVH, and SRH, containing 500 evaluated samples per method. The Dynamic category comprises APH, TRH, and CPH, representing 300 evaluated samples per method. For every evaluation criterion, we compute the accuracy within each group for all fifteen methods and subsequently macro-average the results. The target estimand is defined as the Ontology accuracy minus the Dynamic accuracy. A positive difference therefore indicates higher accuracy on Ontology hallucinations.

The primary percentile bootstrap resamples the 436 video identifiers with replacement, retains all data rows associated with each sampled video, and recomputes the full macro-average. We execute 10,000 draws using seed 20260825 and report the 2.5th and 97.5th percentiles. As a sensitivity check, we additionally resample 100 samples with replacement within each of full categories over 10,000 draws governed by seed 20260826 (Table~\ref{tab:aggregate_bootstrap}).

The within-video OH--DH gap is 19.12 percentage points, close to the full-evaluation gap of 19.53 percentage points (Table~\ref{tab:within_video_comparison}).

\begingroup
\footnotesize
\renewcommand{\arraystretch}{1.10}
\setlength{\tabcolsep}{3.5pt}
\setlength{\LTleft}{0pt plus 1fill}
\setlength{\LTright}{0pt plus 1fill}
\setlength{\LTcapwidth}{\textwidth}
\setlength{\LTpre}{2pt}
\setlength{\LTpost}{2pt}
\begin{table*}[!t]
\centering
\footnotesize
\setlength{\tabcolsep}{2.6pt}
\renewcommand{\arraystretch}{1.06}
\setlength{\tabcolsep}{3.5pt}
\renewcommand{\arraystretch}{1.10}

\resizebox{\textwidth}{!}{%
\begin{tabular}{@{}lccc@{}}
\toprule
\textbf{\mbox{Method}} & \textbf{\mbox{Factual $\Delta$ [95\% CI]}} & \textbf{\mbox{Counterfactual $\Delta$ [95\% CI]}} & \textbf{\mbox{Overall $\Delta$ [95\% CI]}} \\
\midrule
\mbox{Deep Video Discovery \mbox{\cite{zhang2025deepvideodiscovery}}} & $9.67\,[4.63,14.80]^{\dagger}$ & $28.80\,[22.76,34.82]^{\dagger}$ & $36.53\,[30.00,42.91]^{\dagger}$ \\
\mbox{VideoAgent \mbox{\cite{fan2024videoagentmemory}}} & $13.80\,[7.28,20.56]^{\dagger}$ & $3.33\,[-1.11,7.79]$ & $17.87\,[11.52,24.32]^{\dagger}$ \\
\mbox{VideoHV-Agent \mbox{\cite{wang2026think}}} & $2.80\,[-3.16,8.76]$ & $11.40\,[4.08,18.65]^{\dagger}$ & $15.60\,[8.53,22.41]^{\dagger}$ \\
\mbox{Gemini-3-Flash \mbox{\cite{google2025gemini3flash}}} & $-0.13\,[-1.99,1.77]$ & $27.67\,[21.82,33.44]^{\dagger}$ & $26.60\,[20.66,32.40]^{\dagger}$ \\
\mbox{Gemma-4-it \mbox{\cite{gemmateam2026gemma4}}} & $32.33\,[25.46,39.12]^{\dagger}$ & $-0.73\,[-3.91,2.57]$ & $31.93\,[25.15,38.67]^{\dagger}$ \\
\mbox{GPT-5 \mbox{\cite{openai2025gpt5}}} & $14.33\,[8.64,20.13]^{\dagger}$ & $10.27\,[6.13,14.53]^{\dagger}$ & $21.07\,[14.89,27.32]^{\dagger}$ \\
\mbox{InternVL3.5 \mbox{\cite{wang2025internvl35}}} & $2.47\,[-1.85,6.92]$ & $29.00\,[22.28,35.94]^{\dagger}$ & $29.67\,[22.48,36.74]^{\dagger}$ \\
\mbox{LLaVA-NeXT-Video \mbox{\cite{li2024llavanextinterleave}}} & $14.47\,[8.35,20.69]^{\dagger}$ & $-18.73\,[-24.94,-12.53]^{\dagger}$ & $-0.87\,[-3.15,1.17]$ \\
\mbox{Qwen3-Omni-Instruct \mbox{\cite{xu2025qwen3omni}}} & $1.27\,[-0.86,3.49]$ & $16.53\,[9.38,23.58]^{\dagger}$ & $17.73\,[10.58,24.76]^{\dagger}$ \\
\mbox{Qwen3-VL-Instruct \mbox{\cite{bai2025qwen3vl}}} & $9.93\,[4.91,15.18]^{\dagger}$ & $8.07\,[2.48,13.99]^{\dagger}$ & $17.60\,[10.89,24.49]^{\dagger}$ \\
\mbox{Qwen3.6-27B \mbox{\cite{qwenteam2026qwen36}}} & $-0.27\,[-3.27,2.78]$ & $24.27\,[18.73,29.68]^{\dagger}$ & $22.40\,[16.29,28.35]^{\dagger}$ \\
\mbox{EMScore \mbox{\cite{shi2021emscore}}} & $-16.07\,[-22.56,-9.63]^{\dagger}$ & $22.60\,[16.29,28.90]^{\dagger}$ & $4.00\,[0.65,7.24]^{\dagger}$ \\
\mbox{FIFA \mbox{\cite{jing2025fifa}}} & $24.93\,[18.35,31.54]^{\dagger}$ & $-3.93\,[-8.13,0.25]$ & $20.73\,[14.58,27.10]^{\dagger}$ \\
\mbox{PAC-S \mbox{\cite{sarto2023positive}}} & $-7.33\,[-14.36,-0.15]^{\dagger}$ & $15.80\,[8.58,22.91]^{\dagger}$ & $7.33\,[3.96,10.65]^{\dagger}$ \\
\mbox{Owl-Con \mbox{\cite{bansal2024videocon}}} & $-12.00\,[-18.43,-5.57]^{\dagger}$ & $33.47\,[26.92,40.05]^{\dagger}$ & $24.73\,[18.95,30.50]^{\dagger}$ \\*
\bottomrule
\end{tabular}%
}
\caption{Method-Level Performance Differences Between Ontology and Dynamic Hallucinations\equaldash{---}All reported values represent percentage points. Confidence intervals derive from 10,000 video-cluster bootstrap iterations. Positive values designate higher accuracy on the Ontology subset. Negative values correspond to superior performance on the Dynamic subset. A dagger ($^{\dagger}$) marks a 95\% confidence interval strictly excluding zero. LLaVA-NeXT-Video~\cite{li2024llavanextinterleave} is the only method whose Overall interval includes zero, so the Ontology--Dynamic difference is not statistically distinguishable from zero under this bootstrap analysis.}
\label{tab:method_bootstrap}
\end{table*}
\endgroup

\begin{table*}[!t]
\centering
\small
\renewcommand{\arraystretch}{1.10}
\begin{tabular*}{\textwidth}{@{\extracolsep{\fill}}ccccc@{}}
\toprule
\textbf{\mbox{Method}} & \textbf{\mbox{Both correct}} & \textbf{\mbox{Factual only}} & \textbf{\mbox{Counterfactual only}} & \textbf{\mbox{Neither}} \\
\midrule
PAC-S \mbox{\cite{sarto2023positive}} & 7.25 & 40.50 & 49.63 & 2.63 \\
EMScore \mbox{\cite{shi2021emscore}} & 6.50 & 60.13 & 30.63 & 2.75 \\
Owl-Con \mbox{\cite{bansal2024videocon}} & 33.13 & 34.38 & 26.13 & 6.38 \\
FIFA \mbox{\cite{jing2025fifa}} & 34.63 & 9.63 & 54.25 & 1.50 \\
\bottomrule
\end{tabular*}
\caption{Outcome Breakdown Across Selected Thresholds (\%)\equaldash{---}Values denote percentages of evaluated instances, summing to 100\% per method. This decomposition exposes distinct marginal distributions across evaluators. The disparity between isolated and joint successes reveals critical blind spots in fine-grained video temporal understanding.}
\label{tab:detector_pair_outcomes}
\end{table*}

\subsection{Method-Level Intervals}\label{app:method_level_intervals}
Table~\ref{tab:method_bootstrap} details the performance differences for all fifteen evaluated methods. These values derive from the primary video-cluster bootstrap resampling. 

Under the strict evaluation criterion, 14 out of the fifteen methods exhibit intervals entirely above zero. LLaVA-NeXT-Video~\cite{li2024llavanextinterleave} serves as the sole exception to this overarching trend. It records a marginal strict difference of $-0.87$ points, yielding an interval spanning $[-3.15, 1.17]$.

Compared with Ontology samples, EMScore~\cite{shi2021emscore} and Owl-Con~\cite{bansal2024videocon} more frequently accept factual answers on Dynamic samples but less reject counterfactual answers. This pattern indicates a greater tendency to accept candidates on Dynamic samples, including incorrect counterfactual answers.

\subsection{Decomposition of Joint Decision Outcomes}\label{app:joint_decision_decomposition}
Table~\ref{tab:detector_pair_outcomes} reports the distribution of instances across the four decision states. The breakdown reveals clear divergence between isolated and joint performance. Methods such as PAC-S~\cite{sarto2023positive} and EMScore~\cite{shi2021emscore} resolve individual factual or counterfactual assertions, yet their correct classifications rarely coincide on the same instance.

\begin{table*}[!t]
\centering
\noindent\begin{minipage}{\linewidth}
\hrule height 0.7pt
\vspace{2pt}
\centering\textbf{\VidHalLoc{} Example: Action Predicate Hallucination (APH)}\par
\vspace{2pt}
\hrule height 0.7pt
\vspace{3pt}
\includegraphics[width=\linewidth]{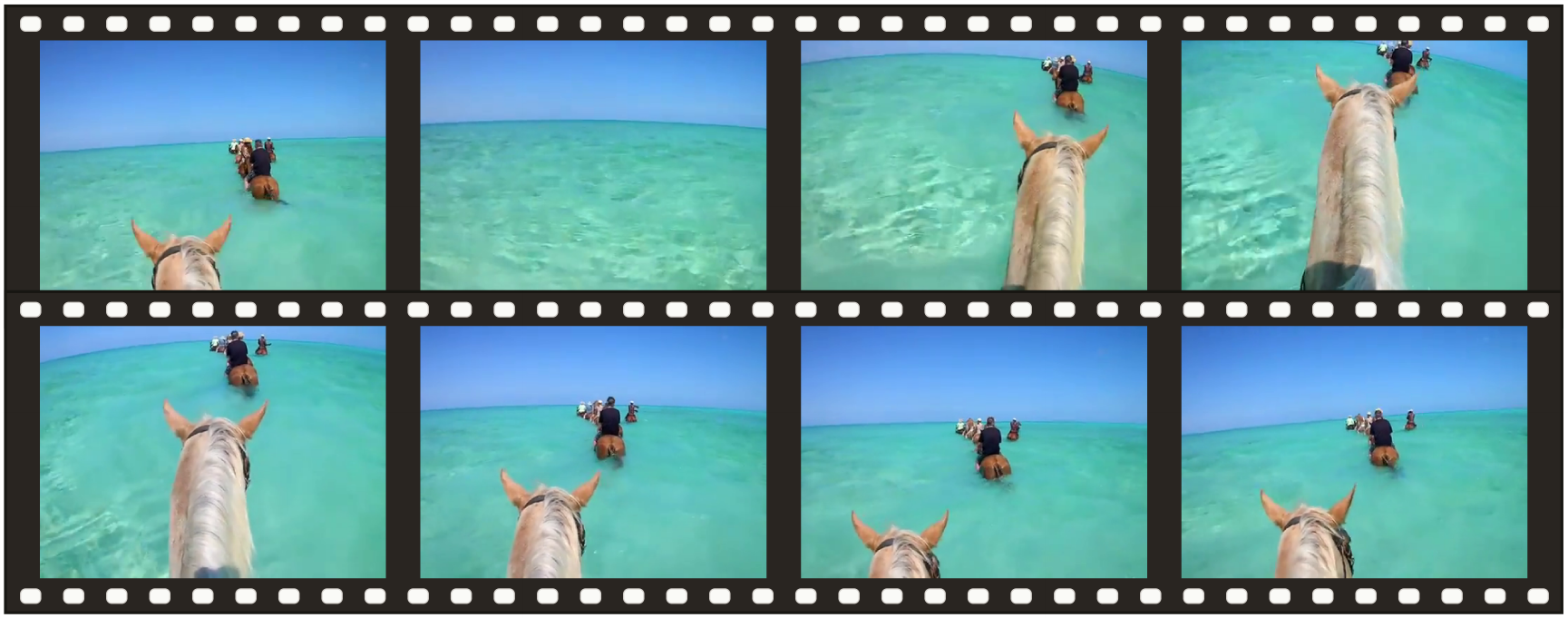}\par
\vspace{2pt}

\begingroup
\footnotesize
\setlength{\tabcolsep}{4pt}
\renewcommand{\arraystretch}{1.08}
\setlength{\emergencystretch}{1em}
\begin{tabular}{@{}>{\RaggedRight\arraybackslash\bfseries}p{0.20\linewidth}
                    >{\justifying\arraybackslash}p{0.725\linewidth}
                    >{\centering\arraybackslash}p{0.03\linewidth}@{}}
Task & Caption & \\
Task prompt & \examplejudgeprompt & \\
Sample question & Identify one fact directly supported by the video and state it in one sentence. & \\
Factual answer & \textcolor{answergreen}{A group of people is riding horses through the shallow ocean water.} & \goodmark \\
Counterfactual answer & \textcolor{answerred}{A group of people is leading horses through the shallow ocean water.} & \badmark \\
Conflict slot & Action predicate: \emph{riding} $\rightarrow$ \emph{leading}. & \\
\end{tabular}

\vspace{3pt}
\hrule height 0.7pt
\endgroup
\end{minipage}

\par\medskip
\centering
\begin{minipage}{\linewidth}
\begingroup
\footnotesize
\setlength{\tabcolsep}{4pt}
\renewcommand{\arraystretch}{1.08}
\setlength{\emergencystretch}{1em}
\begin{tabular*}{\linewidth}{@{\extracolsep{\fill}}lccc@{}}
\toprule
\textbf{\mbox{Method}} & \textbf{\mbox{Factual decision}} & \textbf{\mbox{Counterfactual decision}} & \textbf{\mbox{Strict pair}} \\
\midrule
\multicolumn{4}{@{}l}{\itshape Commercial VLMs} \\
Gemini-3-Flash \mbox{\cite{google2025gemini3flash}} & \textsc{Accept}\,\goodmark & \textsc{Accept}\,\badmark & \badmark \\
GPT-5 \mbox{\cite{openai2025gpt5}} & \textsc{Accept}\,\goodmark & \textsc{Reject}\,\goodmark & \goodmark \\
\addlinespace[1pt]
\multicolumn{4}{@{}l}{\itshape Open source VLMs} \\
LLaVA-NeXT-Video \mbox{\cite{li2024llavanextinterleave}} & \textsc{Accept}\,\goodmark & \textsc{Accept}\,\badmark & \badmark \\
Qwen3-VL-Instruct \mbox{\cite{bai2025qwen3vl}} & \textsc{Accept}\,\goodmark & \textsc{Accept}\,\badmark & \badmark \\
Gemma-4-it \mbox{\cite{gemmateam2026gemma4}} & \textsc{Reject}\,\badmark & \textsc{Reject}\,\goodmark & \badmark \\
InternVL3.5 \mbox{\cite{wang2025internvl35}} & \textsc{Accept}\,\goodmark & \textsc{Accept}\,\badmark & \badmark \\
Qwen3.6-27B \mbox{\cite{qwenteam2026qwen36}} & \textsc{Accept}\,\goodmark & \textsc{Accept}\,\badmark & \badmark \\
Qwen3-Omni-Instruct \mbox{\cite{xu2025qwen3omni}} & \textsc{Accept}\,\goodmark & \textsc{Accept}\,\badmark & \badmark \\
\addlinespace[1pt]
\multicolumn{4}{@{}l}{\itshape Video agents} \\
VideoAgent \mbox{\cite{fan2024videoagentmemory}} & \textsc{Accept}\,\goodmark & \textsc{Accept}\,\badmark & \badmark \\
VideoHV-Agent \mbox{\cite{wang2026think}} & \textsc{Accept}\,\goodmark & \textsc{Accept}\,\badmark & \badmark \\
Deep Video Discovery \mbox{\cite{zhang2025deepvideodiscovery}} & \textsc{Accept}\,\goodmark & \textsc{Reject}\,\goodmark & \goodmark \\
\addlinespace[1pt]
\multicolumn{4}{@{}l}{\itshape Detection methods} \\
PAC-S \mbox{\cite{sarto2023positive}} & \textsc{Accept}\,\goodmark & \textsc{Accept}\,\badmark & \badmark \\
EMScore \mbox{\cite{shi2021emscore}} & \textsc{Accept}\,\goodmark & \textsc{Accept}\,\badmark & \badmark \\
Owl-Con \mbox{\cite{bansal2024videocon}} & \textsc{Accept}\,\goodmark & \textsc{Accept}\,\badmark & \badmark \\
FIFA \mbox{\cite{jing2025fifa}} & \textsc{Accept}\,\goodmark & \textsc{Reject}\,\goodmark & \goodmark \\
\bottomrule
\end{tabular*}
\endgroup
\end{minipage}
\caption{APH example\equaldash{---}GPT-5~\cite{openai2025gpt5}, Deep Video Discovery~\cite{zhang2025deepvideodiscovery}, and FIFA~\cite{jing2025fifa} are the only evaluated methods that accept the factual answer and reject the counterfactual answer at the same time. A green check marks a correct decision, and a red cross marks an incorrect decision.}
\label{tab:aph_qualitative_example}
\end{table*}

\begin{table*}[!t]
\centering
\noindent\begin{minipage}{\linewidth}
\hrule height 0.7pt
\vspace{2pt}
\centering\textbf{\VidHalLoc{} Example: Camera Predicate Hallucination (CPH)}\par
\vspace{2pt}
\hrule height 0.7pt
\vspace{3pt}
\includegraphics[width=\linewidth]{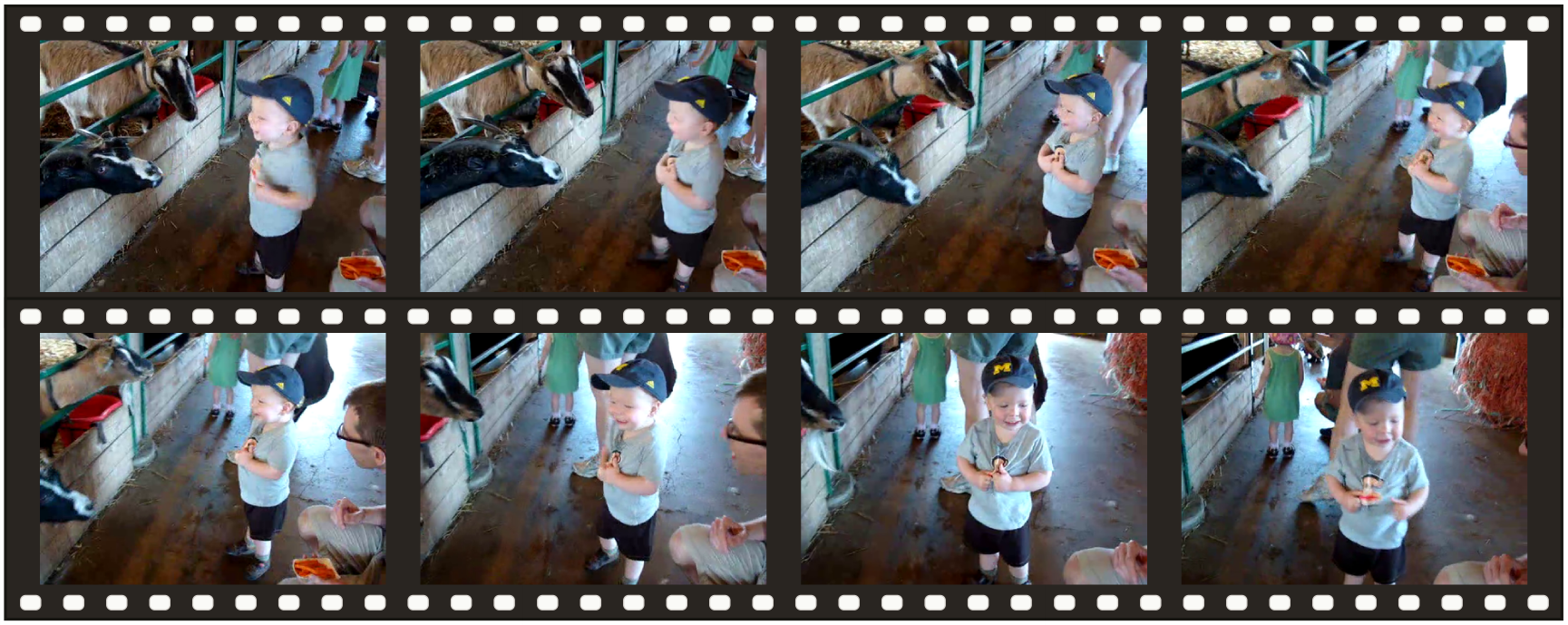}\par
\vspace{2pt}

\begingroup
\footnotesize
\setlength{\tabcolsep}{4pt}
\renewcommand{\arraystretch}{1.08}
\setlength{\emergencystretch}{1em}
\begin{tabular}{@{}>{\RaggedRight\arraybackslash\bfseries}p{0.20\linewidth}
                    >{\justifying\arraybackslash}p{0.725\linewidth}
                    >{\centering\arraybackslash}p{0.03\linewidth}@{}}
Task & Video QA & \\
Task prompt & \examplejudgeprompt & \\
Sample question & What camera or editing change occurs during camera movement between 00:20 and 00:24? & \\
Factual answer & \textcolor{answergreen}{The camera pans right from the goat pen toward the boy.} & \goodmark \\
Counterfactual answer & \textcolor{answerred}{The camera zooms in from the goat pen toward the boy.} & \badmark \\
Conflict slot & Camera predicate: \emph{pans right} $\rightarrow$ \emph{zooms in}. & \\
\end{tabular}

\vspace{3pt}
\hrule height 0.7pt
\endgroup
\end{minipage}

\par\medskip
\centering
\begin{minipage}{\linewidth}
\begingroup
\footnotesize
\setlength{\tabcolsep}{4pt}
\renewcommand{\arraystretch}{1.08}
\setlength{\emergencystretch}{1em}
\begin{tabular*}{\linewidth}{@{\extracolsep{\fill}}lccc@{}}
\toprule
\textbf{\mbox{Method}} & \textbf{\mbox{Factual decision}} & \textbf{\mbox{Counterfactual decision}} & \textbf{\mbox{Strict pair}} \\
\midrule
\multicolumn{4}{@{}l}{\itshape Commercial VLMs} \\
Gemini-3-Flash \mbox{\cite{google2025gemini3flash}} & \textsc{Accept}\,\goodmark & \textsc{Accept}\,\badmark & \badmark \\
GPT-5 \mbox{\cite{openai2025gpt5}} & \textsc{Reject}\,\badmark & \textsc{Accept}\,\badmark & \badmark \\
\addlinespace[1pt]
\multicolumn{4}{@{}l}{\itshape Open source VLMs} \\
LLaVA-NeXT-Video \mbox{\cite{li2024llavanextinterleave}} & \textsc{Reject}\,\badmark & \textsc{Reject}\,\goodmark & \badmark \\
Qwen3-VL-Instruct \mbox{\cite{bai2025qwen3vl}} & \textsc{Accept}\,\goodmark & \textsc{Accept}\,\badmark & \badmark \\
Gemma-4-it \mbox{\cite{gemmateam2026gemma4}} & \textsc{Reject}\,\badmark & \textsc{Reject}\,\goodmark & \badmark \\
InternVL3.5 \mbox{\cite{wang2025internvl35}} & \textsc{Accept}\,\goodmark & \textsc{Accept}\,\badmark & \badmark \\
Qwen3.6-27B \mbox{\cite{qwenteam2026qwen36}} & \textsc{Accept}\,\goodmark & \textsc{Accept}\,\badmark & \badmark \\
Qwen3-Omni-Instruct \mbox{\cite{xu2025qwen3omni}} & \textsc{Accept}\,\goodmark & \textsc{Accept}\,\badmark & \badmark \\
\addlinespace[1pt]
\multicolumn{4}{@{}l}{\itshape Video agents} \\
VideoAgent \mbox{\cite{fan2024videoagentmemory}} & \textsc{Reject}\,\badmark & \textsc{Reject}\,\goodmark & \badmark \\
VideoHV-Agent \mbox{\cite{wang2026think}} & \textsc{Accept}\,\goodmark & \textsc{Accept}\,\badmark & \badmark \\
Deep Video Discovery \mbox{\cite{zhang2025deepvideodiscovery}} & \textsc{Reject}\,\badmark & \textsc{Reject}\,\goodmark & \badmark \\
\addlinespace[1pt]
\multicolumn{4}{@{}l}{\itshape Detection methods} \\
PAC-S \mbox{\cite{sarto2023positive}} & \textsc{Accept}\,\goodmark & \textsc{Accept}\,\badmark & \badmark \\
EMScore \mbox{\cite{shi2021emscore}} & \textsc{Accept}\,\goodmark & \textsc{Accept}\,\badmark & \badmark \\
Owl-Con \mbox{\cite{bansal2024videocon}} & \textsc{Accept}\,\goodmark & \textsc{Accept}\,\badmark & \badmark \\
FIFA \mbox{\cite{jing2025fifa}} & \textsc{Reject}\,\badmark & \textsc{Reject}\,\goodmark & \badmark \\
\bottomrule
\end{tabular*}
\endgroup
\end{minipage}
\caption{CPH example\equaldash{---}None of the evaluated methods correctly resolve both candidates. Instead, most approaches collapse into uniformly accepting or rejecting both options simultaneously. Notably, GPT-5~\cite{openai2025gpt5} completely reverses the required labels. For visual clarity, green checks and red crosses indicate correct and incorrect decisions, respectively.}
\label{tab:cph_qualitative_example}
\end{table*}

\section{Qualitative Reliability Examples}\label{app:qualitative_examples}
\looseness=-1 Tables~\ref{tab:aph_qualitative_example} and~\ref{tab:cph_qualitative_example} present two representative instances from \VidHalLoc{}. The method decisions across these cases expose distinct marginal distributions. Most baselines accept the grounded factual statements while failing to reject the counterfactual modification. This failure pattern underscores the need for improving reliability in video hallucination detection techniques.

\FloatBarrier


\end{document}